\documentclass{article}

\usepackage[main, final]{neurips_2026}

\usepackage[utf8]{inputenc} 
\usepackage[T1]{fontenc}    
\usepackage{hyperref}       
\usepackage{url}            
\usepackage{booktabs}       
\usepackage{amsfonts}       
\usepackage{nicefrac}       
\usepackage{microtype}      
\usepackage{xcolor}         
\usepackage{wrapfig}
\usepackage{graphicx}
\usepackage{multirow} 
\usepackage{duckuments}
\usepackage{tabularx}

\usepackage{amsmath}
\usepackage{wrapfig}
\newcommand{\paragrapht}[1]{\noindent\textbf{#1}}
\definecolor{mypurple}{HTML}{E9E7FD}
\usepackage{xspace}
\newcommand{\ours}{XTrack\xspace}
\usepackage[table]{xcolor}
\title{Repurposing Video Diffusion Transformers for Cross-View Temporal Object Correspondence}

\author{
\textbf{Siyoon Jin}\textsuperscript{\textnormal{1*}} \quad
\textbf{Dahyun Chung}\textsuperscript{\textnormal{1*}} \quad
\textbf{Honggyu An}\textsuperscript{\textnormal{1}} \quad
\textbf{Sangbeom Lim}\textsuperscript{\textnormal{2}} \\[1pt]
\textbf{Wonjin Nam}\textsuperscript{\textnormal{1}} \quad
\textbf{Jaewoo Jung}\textsuperscript{\textnormal{1}} \quad
\textbf{WonJun Moon}\textsuperscript{\textnormal{1}} \quad
\textbf{Seungryong Kim}\textsuperscript{\textnormal{1$\dagger$}} \\[3pt]
{\small \textsuperscript{1}KAIST AI \quad \textsuperscript{2}Korea University} \\
{\small \textsuperscript{*}Equal contribution \quad \textsuperscript{$\dagger$}Corresponding author} \\[2pt]
{\small
\href{https://cvlab-kaist.github.io/XTrack}
{\texttt{https://cvlab-kaist.github.io/XTrack}}
}
}

\begin{document}
\maketitle

\begin{abstract}
    Cross-view object correspondence identifies the same object across views. 
    However, existing methods often rely on frame-wise matching or sequential tracking, which can propagate correspondence errors under appearance changes or occlusions.
    Therefore, we introduce \textbf{cross-view temporal object correspondence}, which requires preserving object identity across both views and time from a query mask. 
    To address this, we propose \textbf{\ours}, a unified framework which repurposes a pretrained video DiT with semantic and spatiotemporal priors for RGB-to-mask latent transport, jointly modeling cross-view matching and temporal tracking.
    However, naive video DiTs do not guarantee accurate cross-view grounding and often attend to semantically similar yet incorrect objects across views. 
    Thus, we introduce focal cross-view attention alignment, hard negative conditioning, and local context crop strategies to ensure reliable grounding. 
    Finally, beyond these architectural enhancements, we also tackle the lack of dense annotation by introducing a pseudo-label curation pipeline that converts sparse Ego-Exo4D annotations into dense supervision to enable effective video-level training.
    Our results demonstrate the effectiveness of \ours as a unified framework, ensuring robust temporal consistency and identity preservation. 
\end{abstract}

\vspace{-10pt}
\section{Introduction}
Cross-view object correspondence~\cite{fu2025objectrelatorenablingcrossviewobject}, which aims to identify the same object across different viewpoints, is a core problem in many applications, including multi-view scene understanding~\cite{grauman2024ego}, video perception~\cite{kang2025egox}, and embodied AI~\cite{kim2019eyes, zhao2023learning}. Prior ego-exo cross-view correspondence methods~\cite{fu2025objectrelatorenablingcrossviewobject, kang2025egox, mur2025mama, hu2025robust} have often relied on \textit{frame-level} matching even with video input. Although some methods incorporate temporal tracking or memory, frame-wise correspondence estimation remains vulnerable to ambiguities that cannot be resolved from individual frames alone.

This limitation is particularly important for real-world video understanding, where object identity must remain consistent across both viewpoints and time.
The same object can appear very differently across viewpoints, and its visibility can change over time due to occlusion, motion blur, or partial observation~\cite{koo2025mv, rajivc2025multi}, resulting correspondence can become ambiguous in the subsequent frames. As a result, correspondences from a single frame become ambiguous and unresolvable in such challenging scenarios but recoverable from temporal context or evidence from the other view. Solving correspondence frame-by-frame therefore risks unstable identity assignment and wrong-object matching over time. This raises the key question: \textit{how can we model object correspondence jointly across viewpoints and time, rather than treating each frame as an independent matching problem?}

\begin{figure*}[t!]
    \centering
    \vspace{-15pt}
\includegraphics[width=1\linewidth]{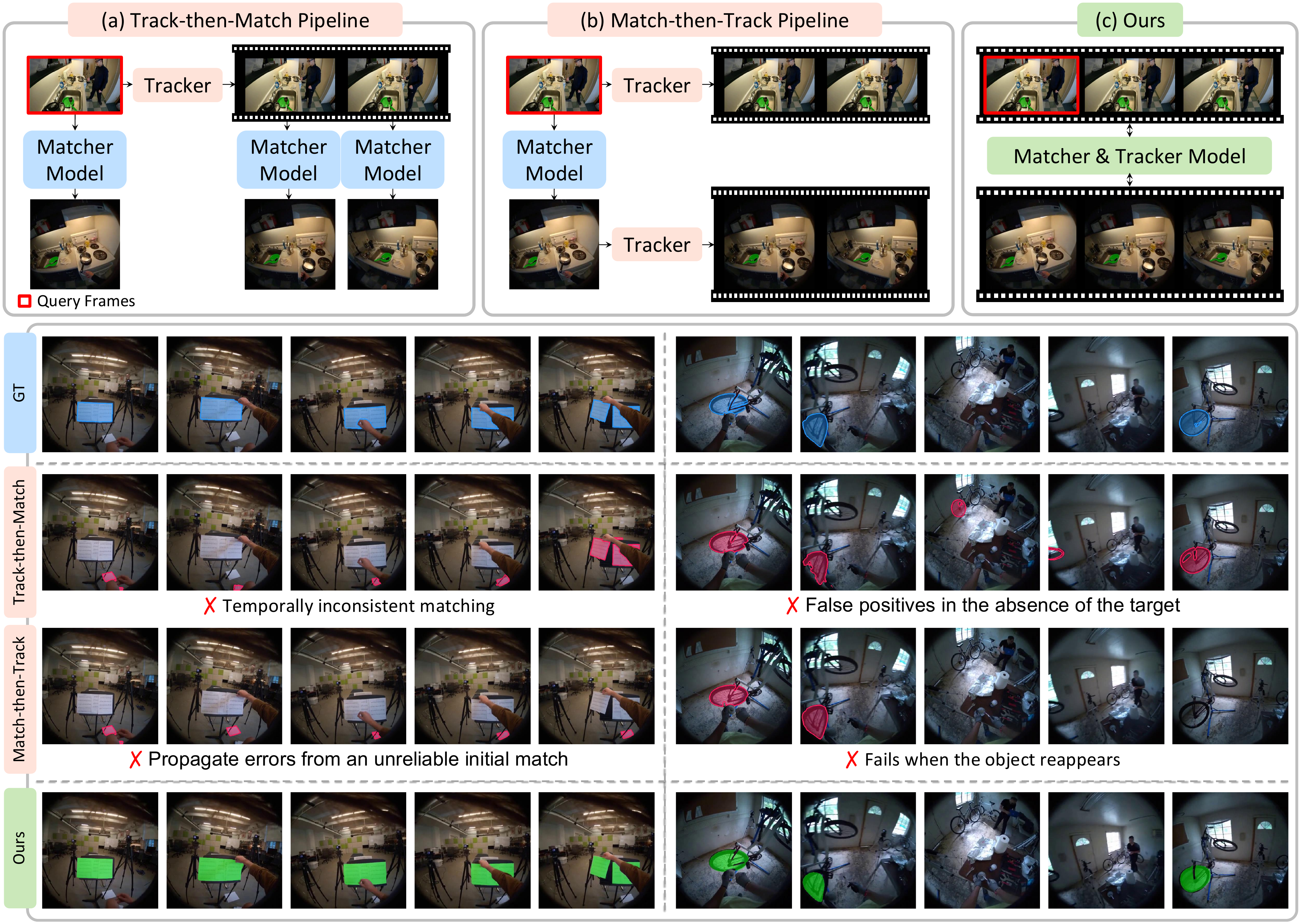}
    \vspace{-15pt}
    \caption{\textbf{Comparison of cross-view temporal object correspondence paradigms.} (Top) (a) propagates the mask in the query view and matches it per frame. (b) starts with initial match, then tracks in each view independently. (c) \ours jointly models cross-view correspondence and temporal dynamics. (Bottom) Comparison of different cross-view correspondence paradigms.}
    \label{fig:paradigm_comparison}
    \vspace{-15pt}
\end{figure*}

We formalize this as \textbf{cross-view temporal object correspondence}. Concretely, given an ego view video, an exo view video, and a first-frame object mask in one view, the goal is to recover temporally coherent object mask sequences in both views. This is more than a temporal extension of frame-level matching. The model must jointly reason about viewpoint variation and temporal dynamics, while handling large appearance changes, temporary disappearance, and ambiguous observations. 

As shown in Fig.~\ref{fig:paradigm_comparison}, we compare three paradigms for this setting. A natural adaptation of frame-level methods is a \textit{track-then-match} pipeline (a), where the first-frame query mask is propagated in query view by a video segmentation model~\cite{ravi2024sam}, and each frame is independently matched to the target view~\cite{mur2025mama}, without considering temporal consistency in the target view. Another extension is a \textit{match-then-track} pipeline (b), 
which estimates an initial match and then propagates in each view independently. However, if the initial match is wrong, the error propagates through the sequence since each view is tracked independently.
More fundamentally, both (a) and (b) decouple cross-view matching from temporal tracking, limiting ability to jointly leverage evidence across views and time.

Motivated by this limitation, we propose a {joint view-time mask prediction formulation}, as illustrated in Fig.~\ref{fig:paradigm_comparison} (c). 
Rather than separating matching from tracking or processing each view independently, this formulation unifies within-view temporal tracking and cross-view matching within a single prediction process. This allows the model to use evidence from one view to resolve uncertainty in the other, and to draw on the full temporal context of both videos throughout prediction.

This formulation naturally raises next question: \textit{what model can support such joint reasoning across views and time?} 
The model must \textit{match} the same object across large viewpoint changes and \textit{track} how objects move and persist over time. 
We argue that these priors are inherently encoded in pretrained \textit{video Diffusion Transformers (DiTs)}~\cite{wan2025wan, yang2024cogvideox, li2024hunyuan}. Through large-scale training, these models develop strong spatiotemporal priors regarding object motion~\cite{nam2025emergent, son2025repurposing}, alongside robust semantic priors that capture how appearances change under varying views, scales, and occlusions~\cite{tang2023emergent, zhang2023tale}.
Leveraging these priors allows us to focus supervision on cross-view temporal mask prediction rather than relearning temporal coherence from scratch.

Motivated by this, we propose \textbf{\ours}, which repurposes a pretrained video DiT~\cite{wan2025wan} under a flow-matching framework~\cite{lipman2022flow} for joint view-time mask prediction from paired RGB videos. Specifically, we feed the ego and exo videos jointly as a diptych~\cite{shin2025largescaletexttoimagemodelinpainting}, that is, a single visual input formed by placing the two views side-by-side, so both views are processed within a shared token space and two mask sequences are inferred within a unified prediction process, keeping both temporal reasoning and cross-view interaction coupled. Since video DiTs do not guarantee accurate cross-view grounding and often attend to semantically similar objects, we further introduce grounding-aware designs that strengthen query-specific grounding. Specifically, we use cross-view attention alignment for query-side grounding, hard negative conditioning to prevent query-agnostic prediction, and local context crops for small objects, preserving their original spatial positions through RoPE~\cite{su2024roformer} alignment.


However, training a video model requires dense and reliable supervision, while Ego-Exo4D~\cite{grauman2024ego} provides only sparse annotations. To address this gap, we build a pseudo-label curation pipeline that converts sparse masks into reliable dense supervision through bidirectional propagation, cross-view verification, and within-view semantic filtering, enabling effective training of our joint video model.

Experiments demonstrate \ours as a unified framework achieving competitive performance with robust temporal consistency and identity preservation. Ablations verify the importance of RGB-to-mask flow prediction, query-specific grounding, local context crops, and pretrained video priors.

\section{Related Works}
\paragrapht{Cross-view Object Correspondence.}
Cross-view object correspondence aims to identify the same object across different views, given a query in one view. It is a core task especially in ego-exo perception~\cite{grauman2024ego}, 
and prior work has explored several directions including view-invariant feature matching~\cite{fu2025objectrelatorenablingcrossviewobject}, mask-level correspondence~\cite{mur2025mama, liao2025domrestablishingcrossviewsegmentation}, and foundation model adaptation~\cite{pan2025v, hu2025robust}.
Despite these advances, many approaches still rely on frame-level matching or sequential correspondence updates, which can propagate errors under ambiguous observations. We address this limitation by formulating cross-view correspondence as a joint view-time prediction problem that unifies cross-view matching and temporal tracking of object identity from the query mask.




\paragrapht{Video Object Segmentation.}
Video object segmentation (VOS)~\cite{ponttuset20182017davischallengevideo, xu2018youtubevoslargescalevideoobject} aims to predict coherent object masks over time from an initial prompt or mask. Prior work preserves temporal identity via memory-based propagation~\cite{oh2019videoobjectsegmentationusing, aksan2019stcnstochastictemporalconvolutional, cheng2022xmemlongtermvideoobject}, decoupled tracking and segmentation~\cite{cheng2023trackingdecoupledvideosegmentation}, query-based mask prediction~\cite{cheng2024puttingobjectvideoobject, yang2021associatingobjectstransformersvideo}, and foundation model adaptation~\cite{ravi2024sam, ding2025sam2longenhancingsam2, yang2024samuraiadaptingsegmentmodel}. Despite this progress, they operate on a single video and are not designed to link object identity across views. 


\paragrapht{Diffusion Models for Perception Tasks.}
Recent approaches~\cite{ke2024repurposing, he2024lotus, nam2023diffusion, tang2023emergent, hedlin2023unsupervised, saxena2023surprising} show that pretrained diffusion models are effective for both generation and perception, with their features and priors repurposed for segmentation~\cite{zhao2023unleashing}, matching~\cite{nam2023diffusion, nam2025emergent}, and depth estimation~\cite{ke2024repurposing, he2024lotus}. This suggests that generative pretraining captures visual structure beyond synthesis. In video, video diffusion models~\cite{blattmann2023stable,  ma2024latte, lin2024open} further learn priors on temporal continuity, object persistence, and appearance variation from large-scale video data. Recent works repurpose these priors for perception tasks such as point tracking~\cite{nam2025emergent, son2025repurposing}, video depth estimation~\cite{hu2025depthcrafter, shao2025learning, zhang2026dvd}, and point map prediction~\cite{jiang2025geo4d, xu2025geometrycrafter}. However, prior diffusion-based perception methods mostly focus on single-view or within-view settings. In contrast, we use a pretrained video DiT for cross-view temporal correspondence to jointly predict object masks across views and over time. 
\section{Preliminaries}

\paragrapht{Flow Matching in Latent Space.}
Given an input $x$, a pretrained VAE~\cite{wan2025wan} encodes it into a latent distribution, where the encoder $\mathcal{E}$ produces $z_0 = \mathcal{E}(x)$, and the decoder $\mathcal{D}$ reconstructs it as $\hat{x} = \mathcal{D}(z_0)$. In this latent space, a diffusion model transports samples from a prior distribution, Gaussian noise $z_1$, to $z_0$. We adopt a flow-matching~\cite{lipman2022flow}, where the model predicts the transport field between two latent endpoints. Given $z_0$ and $z_1$, the intermediate state $z_t$ at time $t \in [0,1]$ is defined as
\begin{equation}
\label{eq:flow_matching}
z_t = (1-t)z_0 + t z_1.
\end{equation}
A neural network $v_\theta (z_t,t,c)$ is trained to predict the transport direction from $z_0$ to $z_1$, given intermediate state $z_t$ at time $t$, conditioned on an input $c$. The model is optimized with:
\begin{equation}
\mathcal{L}_{\mathrm{FM}} = \mathbb{E}_{z_0, z_1, t} \left[ \left\| v_\theta(z_t, t, c) - (z_1 - z_0) \right\|_2^2 \right].
\end{equation}
This formulation learns a continuous transport field between latent states. 

\paragrapht{3D Full Attention of Video Diffusion Transformers.}
A video Diffusion Transformer (DiT)~\cite{wan2025wan} operates on a unified sequence of tokens obtained from its inputs, where all tokens interact through full self-attention. At each layer $l$, their full attention is computed over the entire sequence as:
\begin{equation}
    \text{Attn}(Q_l, K_l, V_l) = A_lV_l, \quad A_l = \text{Softmax}\bigg(\frac{Q_l K_l^\mathsf{{T}}}{\sqrt{d}}\bigg),
    \label{eq:3d_full_attention}
\end{equation}
where $Q_l, K_l$ and $V_l$ are the query, key and value projections at layer $l$, and $d$ is the feature dimension. 
\section{Methods}

\subsection{Problem Formulation and Overview}
Given an ego video $V^\text{ego}$ and an exo video $V^\text{exo}$, along with a first-frame query mask $M_0^\text{q}$ provided in a query view $\text{q}\in\{\text{ego}, \text{exo}\}$, the goal of cross-view temporal object correspondence is to predict temporally coherent mask sequences $\{{M}_t^\text{ego}\}_{t=0}^{T-1}$ and $\{{M}_t^\text{exo}\}_{t=0}^{T-1}$, which denote the predicted object masks in the ego and exo videos over a sequence of length $T$. 
Depending on which view provides the initial query mask, the task is instantiated as \textit{Ego2Exo} or \textit{Exo2Ego}. 
\begin{figure*}[t!]
    \centering
    \vspace{-15pt}
\includegraphics[width=0.95\linewidth]{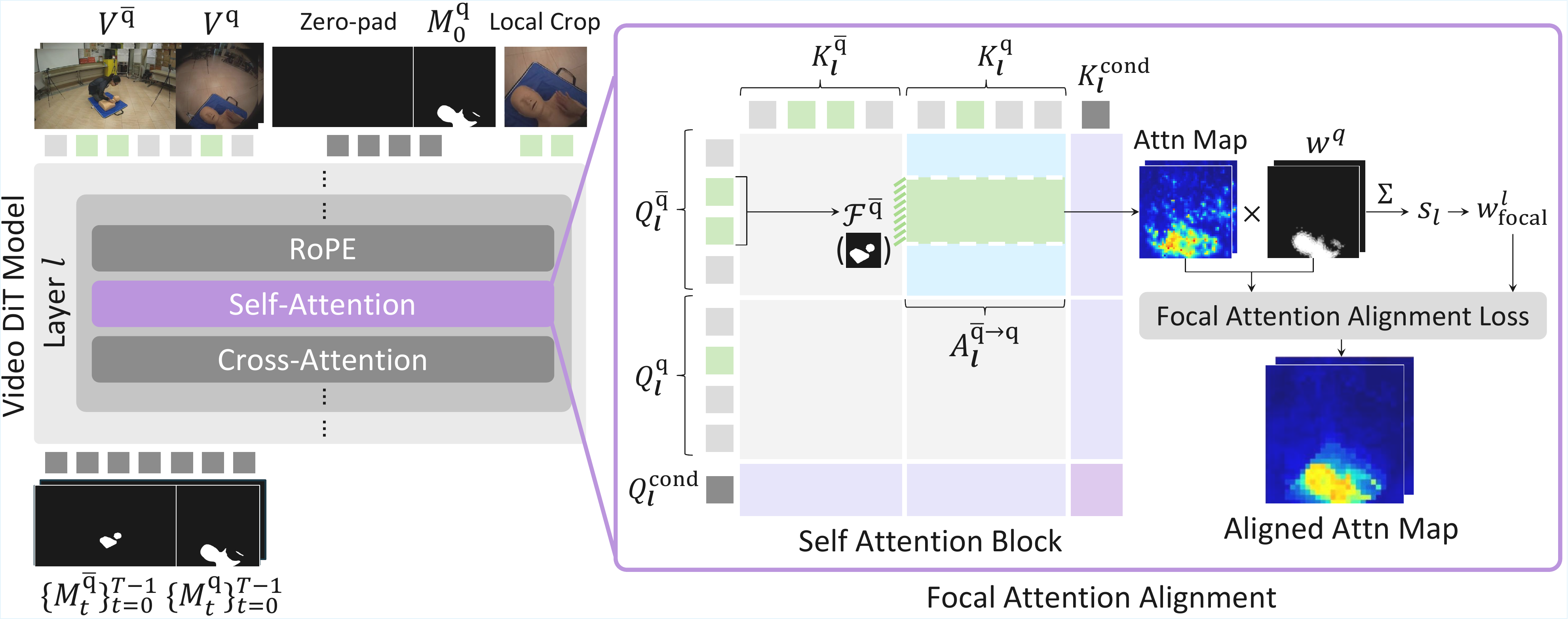}
    \vspace{-10pt}
    \caption{\textbf{Overall architecture of \ours.}}
    \label{fig:overall_architecture}
    \vspace{-15pt}
\end{figure*}

This setting is challenging since cross-view correspondence and temporal matching must be solved simultaneously. The model must identify the same object across views while preserving its identity over time under appearance changes, occlusion and reappearance. Therefore, reliable prediction requires joint reasoning across both views and the temporal dimension, rather than solving matching and tracking independently.
To address this, we propose \textbf{\ours}, which adopts a joint view-time mask prediction formulation using pretrained video DiTs~\cite{wan2025wan}, rather than separating cross-view matching from temporal tracking~\cite{pan2025v}. 
As shown in Fig.~\ref{fig:overall_architecture}, \ours takes $V^{\text{ego}}$ and $V^{\text{exo}}$ along with $M_0^\text{q}$ provided in one view $\text{q}\in\{\text{ego}, \text{exo}\}$ and jointly predicts corresponding mask sequences in both views, leveraging the spatiotemporal priors of video DiTs~\cite{nam2025emergent, son2025repurposing}.
However, vanilla video DiTs do not guarantee query-specific grounding and may attend to semantically similar objects~\cite{tang2023emergent, zhang2023tale}. We thus introduce cross-view object grounding, including focal attention alignment, hard negative conditioning, and local context crops for small objects.
To bridge the gap between the strong temporal prior of video DiTs and their lack of cross-view matching priors, we train \ours in two stages.

\subsection{Cross-view RGB-to-Mask Flow via Video DiT Repurposing}
\label{sec:rgb2mask}
Standard video DiTs~\cite{wan2025wan} are trained to generate temporally coherent videos, implicitly learning to preserve object identity under motion, occlusion, and viewpoint changes. Recent works~\cite{nam2025emergent, son2025repurposing, zhang2023tale, tang2023emergent} show that these priors can be repurposed for tracking tasks, without requiring explicit tracking supervision. These priors provide a strong inductive bias for maintaining object identity, which is essential for consistent mask prediction across views and time.
Our goal, however, is not to synthesize a new RGB video, but to infer object mask sequences grounded in paired ego-exo RGB observations.

Therefore, we repurpose the video DiTs through flow matching by redefining its transport endpoints from noise-to-RGB to RGB-to-mask. In standard video diffusion~\cite{wan2025wan, blattmann2023stable}, the model starts from a Gaussian noise latent that carries no scene information, requiring scene structure and object identity to be constructed during generation. In contrast, starting from the RGB latent provides a source state that already encodes the scene layout, object appearance, and temporal context of the input videos, so the transport focuses on converting the existing visual evidence into mask predictions rather than reconstructing content from scratch. Flow matching provides a natural objective for this transport, as it directly supervises the vector field between the RGB and mask latent endpoints while preserving the generative structure of the pretrained video DiT.
Specifically, following Eq.~\ref{eq:flow_matching}, we set $z_0=z_\text{rgb}$ and $z_1=z_\text{mask}$, where $z_\text{rgb}$ and $z_\text{mask}$ denote the RGB and mask latents of the paired ego-exo inputs. The model is trained to predict the transport direction:
\begin{equation}
    v^*=z_\text{mask} - z_\text{rgb}.
\end{equation}
Conditioned on the paired RGB videos and the one-sided query mask, the model recovers mask latents from RGB latents in a shared latent space. This preserves the video DiT's spatiotemporal and semantic priors while redirecting its generative process toward query-conditioned mask prediction. 


\subsection{Unified Ego-Exo Token Space}
Cross-view temporal correspondence requires the model to resolve ambiguity in one view using evidence from the other. 
If ego and exo videos are processed independently, such cross-view evidence is either unavailable or only introduced through an additional matching stage, making prediction vulnerable when the queried object is occluded or ambiguous in one view.
We therefore process the two views within a unified token space. Specifically, we arrange ego and exo frames as a diptych~\cite{shin2025largescaletexttoimagemodelinpainting}, where the two views are spatially concatenated into a single composite frame. The query mask is provided as conditioning on the query side, while the opposite side receives an empty mask. The video DiT then processes ego-view tokens, exo-view tokens and conditioning tokens jointly within a single spatiotemporal sequence, following  Eq.~\ref{eq:3d_full_attention}. The query and key matrices decompose as:
\begin{equation}
    Q_l = \text{Concat}(Q_l^\text{ego}, Q_l^\text{exo}, Q_l^\text{cond}), \, K_l = \text{Concat}(K_l^\text{ego}, K_l^\text{exo}, K_l^\text{cond}),
\end{equation}
where $Q_l^\text{ego}, Q_l^\text{exo}, Q_l^\text{cond}$ and $K_l^\text{ego}, K_l^\text{exo}, K_l^\text{cond}$ denote the query and key projections of ego, exo, and conditioning tokens at layer $l$. This grouping induces a block structure in the attention matrix, as shown in Fig.~\ref{fig:3d_full_attention}.
We refer to ego-to-ego and exo-to-exo attention as within-view attention, which captures spatial and temporal interactions within each video. In contrast, $A_l^{\text{ego}\rightarrow\text{exo}}$ and $A_l^{\text{exo}\rightarrow\text{ego}}$ denote cross-view attention, where ego queries attend to exo keys and vice versa, to establish object correspondence across views. 
Since full attention is applied globally over both views, tokens in one view can access cross-view evidence at every layer, enabling to resolve ambiguity using cross-view evidence during prediction, rather than relying on a fixed initial match. As a result, cross-view matching and within-view temporal propagation are modeled jointly within a single process.

\subsection{Enhancing Cross-View Object Grounding}
\label{sec:cross-view-obj-grounding}
\paragrapht{Focal Attention Alignment.}
While the unified token space facilitates cross-view interaction, we observe that it may not guarantee instance-level alignment. As shown in Fig.~\ref{supple_fig:attention_visualization}, cross-view attention may diffuse over background regions or distractors~\cite{stracke2025cleandift}, causing the target-side prediction to drift toward the wrong object instance. We therefore directly supervise the cross-view attention blocks using foreground masks. Let $\text{q}\in\{\text{ego, exo}\}$ denote the query-side and $\bar{\text{q}}=\text{q}^c$ denote the target-side. The target-side foreground tokens should attend to the query-side foreground tokens that provide the object evidence. We define the foreground alignment loss as:
\begin{equation}
    \mathcal{L}_\text{pos}^l = -\frac{1}{|\mathcal{F}^{\bar{\text{q}}}|}\sum_{i \in\mathcal{F}^{\bar{\text{q}}}}\log\bigg(\sum_j A_l^{\bar{\text{q}}\rightarrow \text{q}}(i,j)w_j^\text{q} + \epsilon\bigg),
\end{equation}
where $A_l^{\bar{\text{q}}\rightarrow \text{q}}(i,j)$ denotes attention from the $i$-th target token to the $j$-th query token at layer $l$, $\mathcal{F}^{\bar{\text{q}}}$ is the set of target-side foreground tokens, and $w_j^\text{q}$ is the soft foreground weight of the $j$-th query token.  

This loss encourages target-side foreground tokens to attend to the query-side foreground. To avoid disturbing well-grounded cases, we apply a focal weight based on foreground attention confidence,
\begin{equation}
    s_l = \frac{1}{|\mathcal{F}^{\bar{\text{q}}}|} \sum_{i\in\mathcal{F}^{\bar{\text{q}}}}\sum_j A_l^{\bar{\text{q}}\rightarrow \text{q}}(i,j) w_j^\text{q}, \quad w_\text{focal}^l = \lambda_\text{vis}(1-s_l)^\gamma,
\end{equation}
where $\lambda_\text{vis}$ is a foreground visibility factor, $\gamma$ controls how strongly the loss focuses on poorly aligned cases and $\epsilon$ is used for stability. The final alignment loss is :
\begin{equation}
    \mathcal{L}_\text{align} = \frac{1}{L}\sum_{l=1}^L w_\text{focal}^l \mathcal{L}_\text{pos}^l,
\end{equation}
where $L$ is the number of transformer layers. As a result, the loss is strong when cross-view foreground attention is weak, and small when the model already attends to the correct foreground. This formulation covers both Ego2Exo and Exo2Ego by swapping the query and target sides.


\paragrapht{Hard Negative Conditioning.}
Even with attention alignment, the model may produce query-agnostic predictions when similar objects are present, as video diffusion models often attend to semantically similar regions~\cite{zhang2023tale, zhang2024telling}. To enforce instance-level discrimination, we introduce hard negative conditioning. 
Given a correct query mask $M_\text{pos}$, we construct a mismatched query mask $M_\text{neg}$ from another object and compare the predictions with margin loss:
\begin{equation}
    \mathcal{L}_\text{neg} = \max(0, m-\text{MSE}(f_{\bar{\text{q}}}(V^\text{q}, V^{\bar{\text{q}}}, M_\text{pos}), f_{\bar{\text{q}}}(V^\text{q}, V^{\bar{\text{q}}}, M_\text{neg}))),
\end{equation}
where $f_{\bar{\text{q}}}(V^\text{q}, V^{\bar{\text{q}}}, M)$ denotes the target-side prediction given query-side video $V^\text{q}$, target-side video $V^{\bar{\text{q}}}$ and query masks $M$ as inputs and $m>0$ is a margin enforcing predictions under correct and mismatched queries to differ. This loss penalizes similar  predictions under different query conditions, encouraging the model to rely on the query identity rather than generic object saliency.

\paragrapht{Local Context Crop for Small Objects.}
Small objects can occupy only a few tokens in the full-scene latent space, making their appearance cues easy to dilute. To provide stronger object-level evidence, we introduce an object-centered local context crop. The crop is encoded into local tokens and concatenated with the global ego-exo tokens, allowing the model to use both scene-level context and fine-grained object details simultaneously.
To preserve the spatial relationship between the crop and the original scene, we assign RoPE positions based on the crop’s original coordinates in the full frame rather than treating it as a separate image. This keeps the local tokens aligned with the global token layout, allowing them to provide high-resolution object evidence while still referring to the correct scene region. The local crop is treated as an additional conditioning input, alongside the query mask.

\subsection{Training}
\label{sec:training}
\begin{wrapfigure}[7]{r}{0.3\textwidth} 
    \vspace{-15pt}
    \centering
    \includegraphics[width=\linewidth]{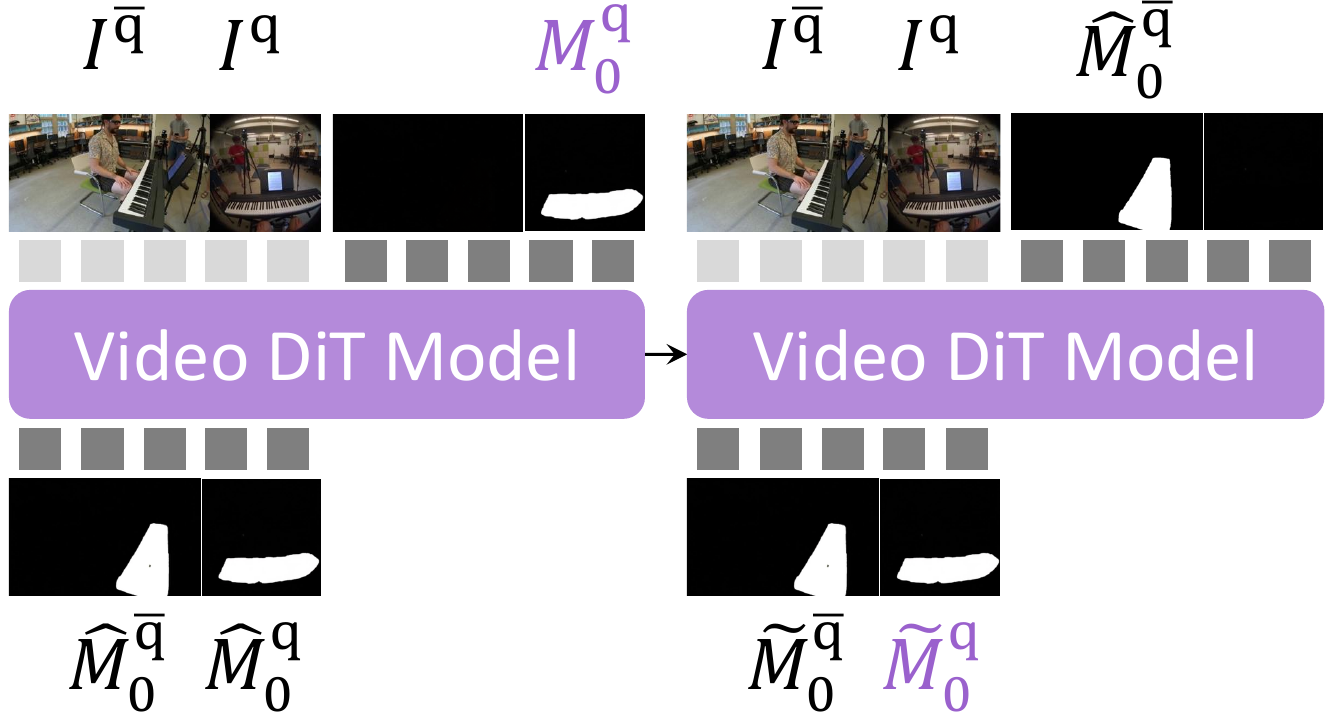}
    \vspace{-15pt}
    \caption{\textbf{2D pretraining with cycle consistency.}}
    \vspace{-15pt}
    \label{fig:2d_cycle_consistency}    
\end{wrapfigure}
Training the full video model requires learning cross-view correspondence and temporal consistency simultaneously. It is challenging since weak correspondence early in training can be carried through the temporal dimension, leading to identity drift or wrong-object propagation. We therefore adopt stage-wise training to stabilize learning. Unlike match-then-track pipelines that first fix a correspondence and then propagate it independently, our model keeps correspondence and propagation coupled within a single video DiT. 

\paragrapht{2D Pretraining with Cycle Consistency.}
We first train on paired ego-exo frame $I^\text{q}$ and $I^{\bar{\text{q}}}$, sampled from videos $V^\text{q}$ and $V^{\bar{\text{q}}}$, allowing the model to focus on cross-view correspondence without temporal variation. 
Given a first-frame query mask $M_0^\text{q}$, the model predicts the corresponding mask in the other view and then maps it back to the original view, 
\begin{equation}
    \hat{M}^{\bar{\text{q}}}_0 = f_{\bar{\text{q}}}(I^\text{q}, I^{\bar{\text{q}}}, M^\text{q}_0), \quad \tilde{M}^\text{q}_0 = f_\text{q}(I^{\bar{\text{q}}}, I^\text{q}, \hat{M}^{\bar{\text{q}}}_0),
\end{equation}
where $f_{\bar{\text{q}}}$ and $f_\text{q}$ denote predictions in the target and query views, and enforce cycle consistency by:
\begin{equation}
    \mathcal{L}_\text{cycle} = ||\tilde{M}^\text{q}_0 - M^\text{q}_0||_2^2,
\end{equation}
which encourages identity-preserving correspondence, as incorrect matches fail to reconstruct the original query mask when mapped back, as shown in Fig.~\ref{fig:2d_cycle_consistency}.
As the cycle constraint implicitly validates the correctness of the predicted correspondence, the model is discouraged from shortcut learning or label memorization.
We apply this objective during 2D pretraining to isolate the cross-view matching problem from temporal dynamics such as motion, occlusion, and drift. By removing temporal complexity, the model is encouraged to learn robust image-level correspondence, which provides a strong initialization for subsequent video-level fine-tuning.


\paragrapht{Video-level Fine-tuning.}
We then fine-tune on video sequences to extend the image-level matching prior to temporal prediction, utilizing dense pseudo-labels generated by our curation pipeline, as detailed in Sec.~\ref{sec:dataset_curation}. This encourages the model to learn to maintain object identity under motion, occlusion, and reappearance while preserving cross-view grounding. At inference, the model does not separate matching and tracking. Instead, it jointly predicts both views over time, allowing cross-view correspondence and temporal propagation to be updated together within the unified token space.



\section{Pseudo-Label Curation Pipeline}
\label{sec:dataset_curation}
\begin{figure*}[t!]
    \centering
    \vspace{-15pt}
\includegraphics[width=0.9\linewidth]{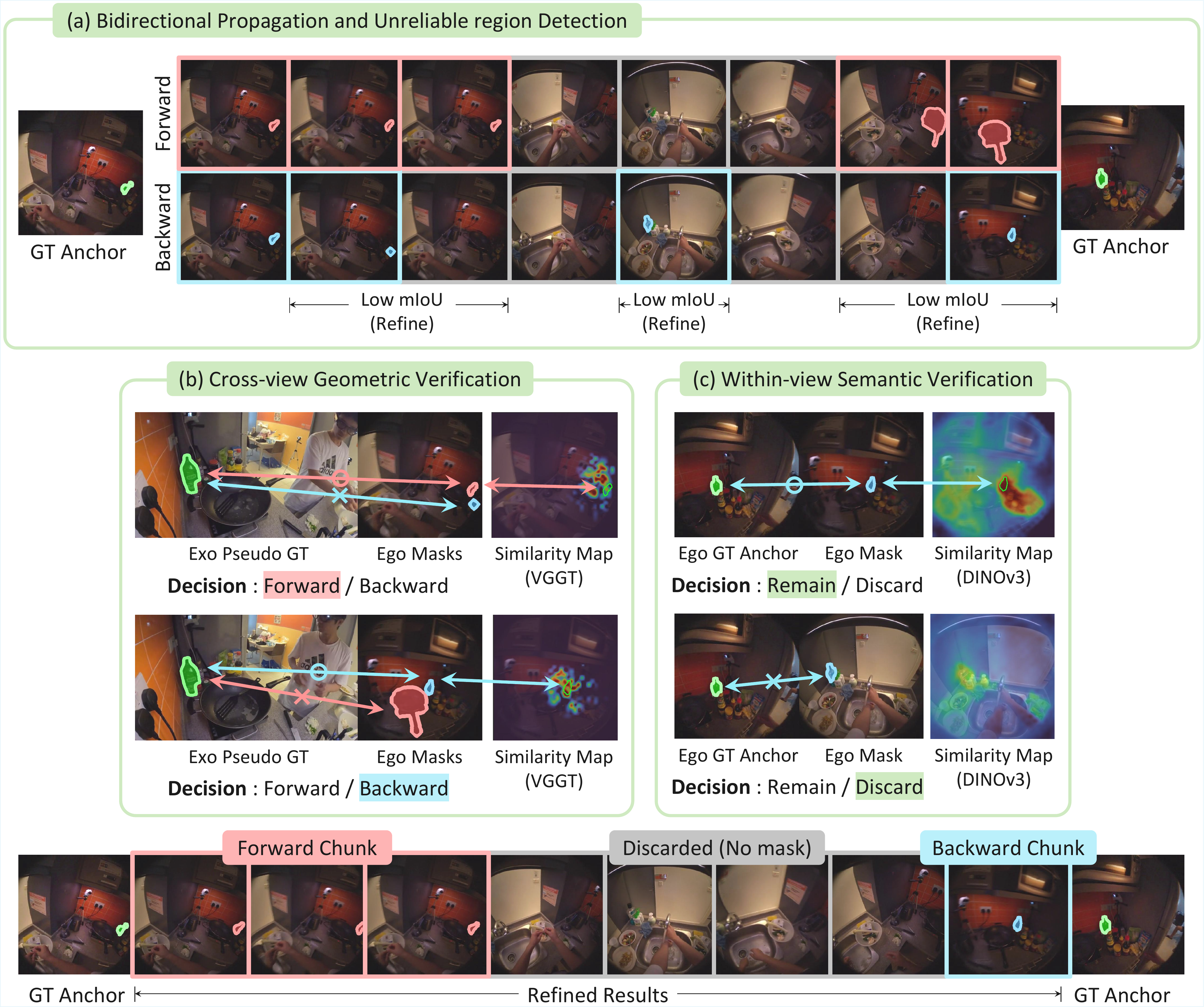}
    \caption{\textbf{Pseudo-label curation pipeline.} (a) performs bidirectional propagation from anchor frames and identifies unreliable regions, which are refined by (b) cross-view geometric verification and further filtered by (c) within-view semantic verification.}
    \vspace{-15pt}
    \label{fig:dataset_curation_pipeline}
\end{figure*}

\paragrapht{Motivation.}
Ego-Exo4D~\cite{grauman2024ego} provides videos at 30 FPS but mask annotations only at 1 FPS, making temporally dense supervision unavailable. A straightforward solution is to propagate sparse annotations using a propagation model~\cite{ravi2024sam}. 
However, we find that SAM2 alone is unreliable in ego–exo cross-view videos, especially in the ego view where occlusions and large viewpoint changes often cause tracking failures. The key challenge is therefore to identify which propagated masks are reliable enough for supervision. To address this, we propose an automatic curation pipeline that leverages cross-view information from the exo view to detect and refine unreliable ego-view pseudo labels, combining geometric and semantic cues from foundation models~\cite{simeoni2025dinov3,wang2025vggt}.

\paragrapht{Bidirectional Mask Propagation and Refinement Region Selection.}
To obtain an internal estimate of reliability, we perform bidirectional propagation from two annotated anchor frames, one forward and one backward, as shown in Fig.~\ref{fig:dataset_curation_pipeline} (a). If propagation is stable, both paths should produce consistent masks in the intermediate region. In practice, however, they often diverge due to occlusion, rapid viewpoint changes, and hand-object interactions. We use this forward-backward inconsistency as a reliability signal. Specifically, we compute the mIoU between the forward and backward masks within each chunk, and treat low-agreement chunks as refinement targets. 

\paragrapht{Cross-view Geometric Verification.}
\label{sec:dataset-geometric}
Ego-view propagation often fails when the object reappears after occlusion, producing inconsistent forward and backward candidates. To resolve this ambiguity, we use exo view propagated masks as pseudo ground truth, since exo videos in Ego-Exo4D~\cite{grauman2024ego} are captured by static cameras, providing more stable masks than ego views. Since the exo reference is captured at the same time, it provides more direct evidence than temporally-distant ego anchors. A natural alternative is explicit geometric triangulation, but it is unreliable due to fisheye distortion in ego cameras. Instead, we use geometry-aware VGGT features~\cite{wang2025vggt} learned from large-scale pretraining, which provide robust cross-view correspondence without requiring undistorted imagery. For each unreliable frame, we extract VGGT features from the forward/backward ego-mask candidates and the exo reference, and compute similarity maps between each ego candidate and the exo reference. As shown in Fig.~\ref{fig:dataset_curation_pipeline} (b), we select the candidate whose similarity exceeds a predefined threshold. 

\paragrapht{Within-view Semantic Verification.}
Cross-view geometric verification resolves ambiguity using cross-view evidence. To further ensure identity consistency, we complement it with a within-view semantic verification stage.
We verify whether the selected candidate remains consistent with a nearby within-view anchor. Since geometric matching is unreliable within a single view, we instead use DINOv3~\cite{simeoni2025dinov3} features to measure semantic similarity. Specifically, we extract DINOv3 features from the candidate, as in the geometric verification, and the neighboring within-view anchor ground-truth, and retain the candidate only if the similarity exceeds a predefined threshold, as shown in Fig.~\ref{fig:dataset_curation_pipeline} (c). 

After these steps, each unreliable chunk is either corrected by cross-view verification or removed by semantic filtering. The final dense supervision retains only masks satisfying both geometric and semantic criteria, producing cleaner and more temporally consistent tracks. Details are in Appendix~\ref{supple_sec:curation_details}.


\section{Experiments}
\begin{figure*}[t!]
    \centering
    \vspace{-15pt}
\includegraphics[width=1\linewidth]{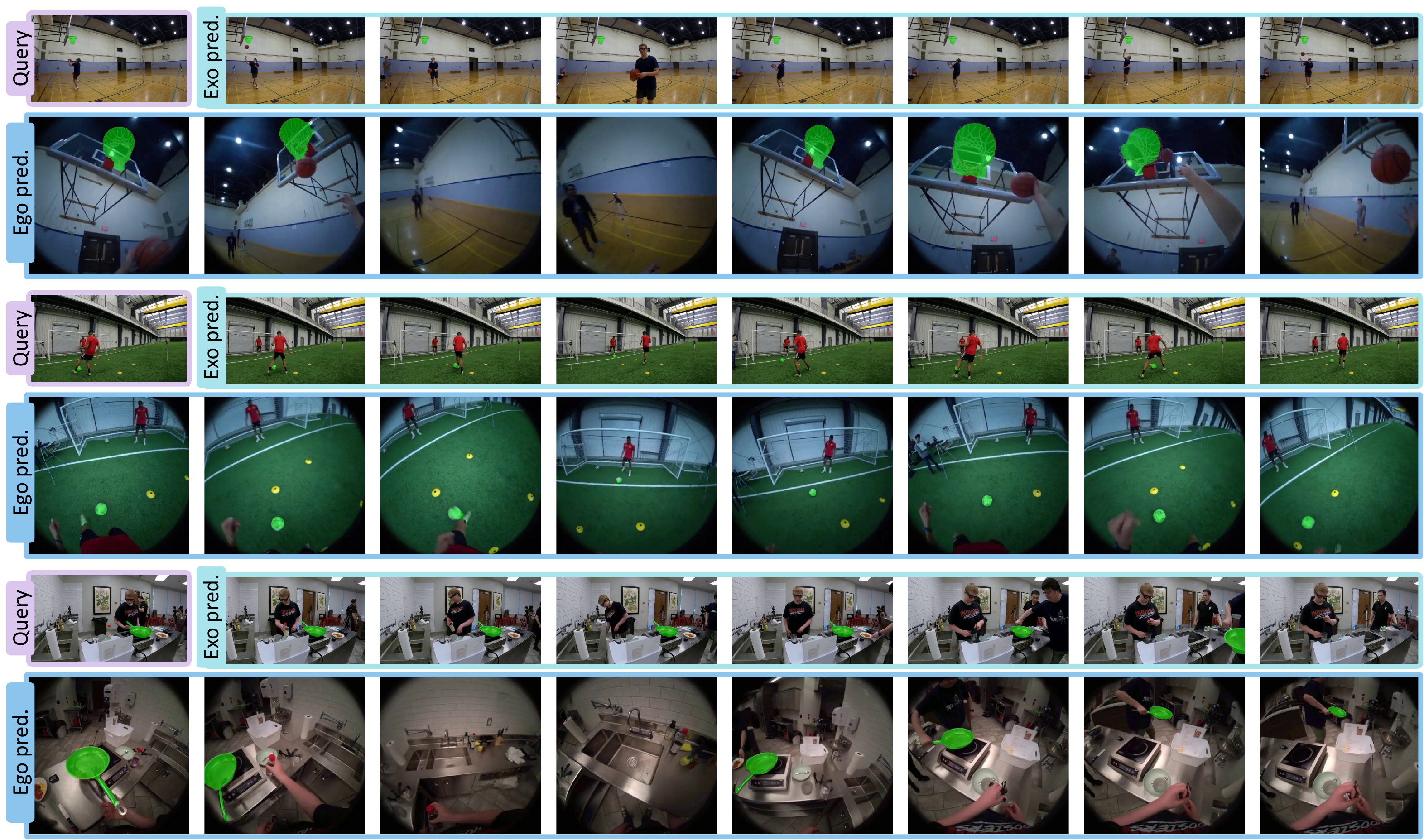}
    \vspace{-15pt}
    \caption{\textbf{Qualitative results on Ego-Exo4D~\cite{grauman2024ego}.} Results are shown for the Exo2Ego setting. Additional Exo2Ego results are provided in Fig.~\ref{supple_fig:additional_qual}, and the opposite setting is shown in Fig.~\ref{supple_fig:additional_qual_ego2exo}.}
    \vspace{-15pt}
    \label{fig:main_qualitative_results}
\end{figure*}

\subsection{Experimental Setup}
\label{sec:experimental_setup}
\paragrapht{Dataset.}
We evaluate on a subset of the Ego-Exo4D~\cite{grauman2024ego} validation split, where dense annotations are constructed using our curation pipeline and additional human verification. Each sample is defined for a specific (ego view, exo view, object) tuple, resulting in a challenging benchmark that emphasizes both cross-view correspondence and long-term tracking stability. Details are in Appendix~\ref{supple_sec:curation_details:dataset_examples} and \ref{supple_sec:evaluation}.


\paragrapht{Metrics.}
We evaluate segmentation quality using region similarity ($\mathcal{J}$), contour accuracy ($\mathcal{F}$), and their average ($\mathcal{J}\&\mathcal{F}$)~\cite{ponttuset20182017davischallengevideo}. We report location error ($\mathrm{LE}$), and visibility accuracy ($\mathrm{VA}$) following Ego-Exo4D~\cite{grauman2024ego}. For temporal consistency, we use mean conditional recall metric (mR @ IoU), which measures tracking retention after successful initial match~\cite{lin2015microsoftcococommonobjects, Everingham15, caelles2017oneshotvideoobjectsegmentation}. Details are in Appendix~\ref{supple_sec:evaluation}.


\paragrapht{Baselines.}
We compare three official baselines~\cite{grauman2024ego} and four models~\cite{zhang2024psalm, fu2025objectrelatorenablingcrossviewobject, mur2025mama, pan2025v} under two settings. In \textit{Track-then-Match}, SAM2~\cite{ravi2024sam} propagates a query mask along the query view, and each propagated mask is used as a query for frame-by-frame prediction. In \textit{Match-then-Track}, the matching model predicts the single-frame target mask, after which SAM2 propagates both views independently. 

\subsection{Main Results}

Tab.~\ref{tab:main_quan_unified} compares \ours against prior cross-view correspondence models evaluated under \textit{Track-then-Match} (T$\rightarrow$M) and \textit{Match-then-Track} (M$\rightarrow$T) pipelines. 
These baselines are fundamentally constrained by their decoupled matching and tracking mechanisms. 
Specifically, the T$\rightarrow$M pipeline relies heavily on frame-level matching. Since target-side masks are predicted independently without temporal context, it suffers from severe temporal inconsistency, yielding notably low VA and mR@IoU scores. 
Conversely, M$\rightarrow$T uses a strong SAM2~\citep{ravi2024sam}-assisted tracking baseline that ensures robust within-view propagation. 
We find that establishing an initial match and subsequently tracking it with a powerful model like SAM2 generally provides greater overall stability than the T$\rightarrow$M approach. While SAM2 benefits from large-scale annotation, which guarantees high mask quality given a reliable prompt and naturally translates to strong performance on mask-quality-related metrics, our unified \ours framework still achieves highly competitive performance, particularly in VA and mR@IoU, by jointly reasoning across views and time within a unified token space.
We do, however, observe a trade-off regarding per-frame spatial precision ($\mathcal{J}\&\mathcal{F}$). 
We attribute this to the highly compressed latent space of the video DiT, where fine-grained boundaries for small objects can be diluted during decoding.
As a result, the model accurately localizes the object's presence (maintaining high VA) but occasionally struggles to delineate its precise contours (visualization in Appendix Fig.~\ref{supple_fig:v2sam_comparison}).
Nevertheless, \ours offers a key architectural advantage in replacing complex pipelines that stitch multiple foundation models with a unified framework with superior tracking capabilities, as evidenced by our high $\text{VA}$ and $\text{mR@IoU}$. 
Fig.~\ref{fig:main_qualitative_results} illustrates our qualitative results.

\begin{table*}
\centering
\vspace{-15pt}
\setlength{\tabcolsep}{3pt}
\renewcommand{\arraystretch}{0.92}
\caption{
\textbf{Quantitative comparison on Ego-Exo4D~\cite{grauman2024ego} under different pipelines.}
Track-then-Match (T$\rightarrow$M) and Match-then-Track (M$\rightarrow$T) are reported.
\textbf{\ours} performs joint prediction independent of pipeline choice. $\text{N/A}$ indicates results unavailable due to missing checkpoints.
}
\label{tab:main_quan_unified}
\resizebox{1\linewidth}{!}{
\begin{tabular}{l c | cccccc | cccccc}
\toprule
\multirow{2}{*}{Method} & \multirow{2}{*}{Pipeline} 
& \multicolumn{6}{c|}{\textbf{Ego2Exo} (Ego as Query)} 
& \multicolumn{6}{c}{\textbf{Exo2Ego} (Exo as Query)} \\
& 
& $\mathcal{J} (\uparrow)$ & $\mathcal{F} (\uparrow)$ & $\mathcal{J\&F} (\uparrow)$ & LE ($\downarrow$) & VA ($\uparrow)$ & mR@IoU ($\uparrow$) 
& $\mathcal{J} (\uparrow)$ & $\mathcal{F} (\uparrow)$ & $\mathcal{J\&F} (\uparrow)$ & LE ($\downarrow$) & VA ($\uparrow)$ & mR@IoU ($\uparrow$) \\
\midrule

\multicolumn{14}{c}{\textit{Track-then-Match (T$\rightarrow$M)}} \\
\midrule
PSALM (Zero-shot)~\cite{zhang2024psalm} & T$\rightarrow$M & 0.126 & 0.155 & 0.140 & 0.221 & 0.480 & 0.633 & 0.149 & 0.180 & 0.165 & 0.168 & 0.601 & 0.502 \\
XSegTx~\cite{grauman2024ego} & T$\rightarrow$M & 0.056 & 0.108 & 0.082 & 0.177 & 0.448 & 0.800 & 0.357 & 0.389 & 0.373 & 0.089 & \underline{0.809} & 0.677 \\
XView - XMem~\cite{grauman2024ego} & T$\rightarrow$M & 0.152 & 0.211 & 0.182 & 0.340 & 0.366 & 0.692 & 0.309 & 0.356 & 0.333 & 0.251 & 0.403 & 0.773 \\
XView - XMem + XSegTx~\cite{grauman2024ego} & T$\rightarrow$M & 0.327 & 0.413 & 0.370 & 0.182 & 0.423 & 0.799 & 0.412 & 0.460 & 0.436 & 0.128 & 0.463 & 0.757 \\
ObjectRelator~\cite{fu2025objectrelatorenablingcrossviewobject} & T$\rightarrow$M & 0.366 & 0.440 & 0.403 & 0.139 & 0.446 & 0.796 & 0.444 & 0.497 & 0.470 & \underline{0.070} & 0.495 & 0.695 \\
O-MaMa~\cite{mur2025mama} & T$\rightarrow$M & $\text{N/A}$ & $\text{N/A}$ & $\text{N/A}$ & $\text{N/A}$ & $\text{N/A}$ & $\text{N/A}$ & 0.314 & 0.365 & 0.340 & 0.115 & 0.496 & 0.555 \\
V$^2$-SAM~\cite{pan2025v} & T$\rightarrow$M & 0.436 & 0.527 & 0.482 & \underline{0.072} & 0.406 & 0.696 & 0.483 & 0.527 & 0.505 & \textbf{0.064} & 0.490 & 0.683 \\

\midrule
\multicolumn{14}{c}{\textit{Match-then-Track (M$\rightarrow$T)}} \\
\midrule
PSALM (Zero-shot)~\cite{zhang2024psalm} & M$\rightarrow$T & 0.146 & 0.183 & 0.164 & 0.139 & 0.502 & 0.938 & 0.203 & 0.237 & 0.220 & 0.167 & 0.596 & 0.839 \\
XSegTx~\cite{grauman2024ego} & M$\rightarrow$T & 0.107 & 0.156 & 0.132 & 0.116 & 0.570 & 0.722 & 0.461 & 0.480 & 0.471 & 0.109 & 0.699 & 0.884 \\
XView - XMem~\cite{grauman2024ego} & M$\rightarrow$T & 0.175 & 0.219 & 0.197 & 0.434 & 0.640 & 0.909 & 0.357 & 0.387 & 0.372 & 0.347 & 0.723 & 0.887 \\
XView - XMem + XSegTx~\cite{grauman2024ego} & M$\rightarrow$T & 0.375 & 0.440 & 0.407 & 0.293 & \underline{0.705} & 0.951 & 0.451 & 0.482 & 0.466 & 0.207 & 0.736 & 0.901 \\
ObjectRelator~\cite{fu2025objectrelatorenablingcrossviewobject} & M$\rightarrow$T & 0.490 & 0.561 & 0.525 & \textbf{0.059} & 0.626 & \textbf{0.969} & 0.489 & \underline{0.529} & 0.509 & 0.098 & 0.716 & \underline{0.903} \\
O-MaMa~\cite{mur2025mama} & M$\rightarrow$T & $\text{N/A}$ & $\text{N/A}$ & $\text{N/A}$ & $\text{N/A}$ & $\text{N/A}$ & $\text{N/A}$ & 0.343 & 0.382 & 0.362 & 0.143 & 0.656 & 0.861 \\
V$^2$-SAM~\cite{pan2025v} & M$\rightarrow$T & \textbf{0.515} & \textbf{0.574} & \textbf{0.545} & 0.100 & 0.628 & 0.954 & \textbf{0.527} & \textbf{0.554} & \textbf{0.540} & 0.104 & 0.691 & 0.898 \\

\midrule
\cellcolor{mypurple}\textbf{\ours (Ours)} & \cellcolor{mypurple}\textbf{Unified} & \cellcolor{mypurple}0.406 & \cellcolor{mypurple}0.538  & \cellcolor{mypurple}0.472 & \cellcolor{mypurple}0.115  & \cellcolor{mypurple}\textbf{0.900} & \cellcolor{mypurple}\underline{0.954}  & \cellcolor{mypurple} \underline{0.490} & \cellcolor{mypurple}{0.528} & \cellcolor{mypurple}\underline{0.509} & \cellcolor{mypurple}0.090 & \cellcolor{mypurple}\textbf{0.814} & \cellcolor{mypurple}\textbf{0.911} \\
\bottomrule
\end{tabular}
               }
\vspace{-15pt}
\end{table*}

\subsection{Ablation Studies}




\begin{wraptable}[6]{r}{0.5\textwidth}  
\centering
\setlength{\tabcolsep}{1.5pt}
\vspace{-18pt}
\caption{\textbf{Ablation studies.}}
\label{tab:ablation}
\resizebox{1\linewidth}{!}{
\begin{tabular}{l|l|ccc|ccc}
\toprule
& \multirow{2}{*}{Method} 
& \multicolumn{6}{c}{\textbf{Exo2Ego} (Exo as Query)} \\
& 
& Ego $\mathcal{J\&F}$ & Ego LE  & Ego VA & Exo $\mathcal{J\&F}$ & Exo LE  & Exo VA\\
\midrule

(I) & Baseline & 0.303 & 0.134 & 0.674 & 0.415 & 0.143 & 0.859 \\
(II) & (I) + $\mathcal{L}_\text{align}$ & 0.332 & 0.129 & 0.755 & 0.514 & 0.161 & \underline{0.906} \\
(III) & (II) + Hard Neg. & \underline{0.431} & \underline{0.106} & \underline{0.776} & \underline{0.579} & \underline{0.073} & 0.882  \\    
\cellcolor{mypurple}(IV) & \cellcolor{mypurple}(III) + Local Crop (\textbf{Ours}) & \cellcolor{mypurple}\textbf{0.509} & \cellcolor{mypurple}\textbf{0.090} & \cellcolor{mypurple}\textbf{0.814} & \cellcolor{mypurple}\textbf{0.601} & \cellcolor{mypurple}\textbf{0.060} & \cellcolor{mypurple}\textbf{0.908} \\
(V) & (IV) w/o pretrained weight & 0.054 & 0.189 & 0.647 & 0.025 & 0.243 & 0.845 \\

\bottomrule
\end{tabular}
\vspace{-15pt}
}
\end{wraptable}
Tab.~\ref{tab:ablation} isolates the effect of each design choice. (I) is baseline, which uses only the unified RGB-to-mask joint prediction framework without additional grounding-specific designs. Adding $\mathcal{L}_\text{align}$ in (II) improves both query-side and target-side prediction, showing that cross-view attention supervision benefits joint mask sequence prediction rather than only the target view. (III) adds hard negative conditioning, and yields a large gain in mask quality, indicating that instance-level discrimination is crucial for preventing confusion with similar objects. (IV) further adds local context crops and achieves superior overall performance, indicating that fine-grained object evidence complements cross-view grounding. 
Finally, removing pretrained weights in (V) causes a clear drop, confirming the importance of pretrained video DiT priors.

\section{Conclusion}
We introduce cross-view temporal object correspondence that requires preserving object identity jointly across views and over time from a single query mask. Unlike prior methods that rely on decoupled frame-level matching, we propose joint view-time mask prediction, a unified formulation that couples cross-view matching and temporal tracking within a video DiT. We instantiate this through \ours, which repurposes a video DiT to perform RGB-to-mask latent transport over paired ego-exo videos. By leveraging the spatiotemporal priors of the video DiT alongside targeted cross-view object grounding mechanisms, our framework ensures query-conditioned mask predictions. We further develop a pseudo-label curation pipeline that converts sparse annotations into reliable dense supervision for effective training of \ours. Experiments show that \ours produces temporally consistent and identity-preserving mask sequences across views, with comparable mask quality.

\newpage
\clearpage

{
    \small
    \bibliographystyle{plain} 
    \bibliography{egbib}
}
\medskip

\small





\newpage
\appendix
\section*{\LARGE Appendix}
\noindent This appendix provides supplementary material to support the main paper.

\noindent\textbf{Table of Contents}

\vspace{0.5em}
\noindent\textbf{A\quad Challenges of Cross-View Object Correspondence} \dotfill \pageref{supple_sec:challenges}\\[2pt]

\vspace{0.5em}
\noindent\textbf{B\quad \ours Details} \dotfill \pageref{supple_sec:impl_details}\\[2pt]
\noindent\hspace*{1.5em}B.1\quad Architectural Details \dotfill \pageref{supple_sec:impl_details:arch_details}\\[2pt]
\noindent\hspace*{1.5em}B.2\quad Cross-View Object Grounding Details \dotfill \pageref{supple_sec:impl_details:grounding_details}\\[2pt]
\noindent\hspace*{1.5em}B.3\quad Additional Training Strategy \dotfill \pageref{supple_sec:impl_details:training}\\[2pt]
\noindent\hspace*{1.5em}B.4\quad Implementation Details \dotfill \pageref{supple_sec:impl_details:impl}\\[2pt]

\vspace{0.5em}
\noindent\textbf{C\quad Pseudo-Label Curation Pipeline Details} \dotfill \pageref{supple_sec:curation_details}\\[2pt]
\noindent\hspace*{1.5em}C.1\quad Bidirectional Propagation Details \dotfill \pageref{supple_sec:curation_details:propagation_details}\\[2pt]
\noindent\hspace*{1.5em}C.2\quad Unreliable Chunk Selection Details\dotfill \pageref{supple_sec:curation_details:chunk_selection_details}\\[2pt]
\noindent\hspace*{1.5em}C.3\quad Cross-view Geometric Verification Details \dotfill \pageref{supple_sec:curation_details:within-view_details}\\[2pt]
\noindent\hspace*{1.5em}C.4\quad Within-view Semantic Verification Details\dotfill \pageref{supple_sec:curation_details:within-view_details}\\[2pt]
\noindent\hspace*{1.5em}C.5\quad 
Dataset Examples and Statistics\dotfill \pageref{supple_sec:curation_details:dataset_examples}\\[2pt]

\vspace{0.5em}
\noindent\textbf{D\quad Evaluation} \dotfill \pageref{supple_sec:evaluation}\\[2pt]
\noindent\hspace*{1.5em}D.1\quad Evaluation Dataset\dotfill \pageref{supple_sec:evaluation:evaluation_dataset}\\[2pt]
\noindent\hspace*{1.5em}D.2\quad Evaluation Metric\dotfill \pageref{supple_sec:evaluation:evaluation_metric}\\[2pt]

\vspace{0.5em}
\noindent\textbf{E\quad Analyses} \dotfill \pageref{supple_sec:analyses}\\[2pt]
\noindent\hspace*{1.5em}E.1\quad Analysis for Design Choice \dotfill \pageref{supple_sec:analyses:analysis_design_choice}\\[2pt]
\noindent\hspace*{1.5em}E.2\quad Analysis for Attention Alignment \dotfill \pageref{supple_sec:analyses:attention}\\[2pt]
\noindent\hspace*{1.5em}E.3\quad Effect of Local Context Crops on Small Objects \dotfill \pageref{supple_sec:analyses:small_object_analysis}\\[2pt]

\vspace{0.5em}
\noindent\textbf{F\quad Additional Results} \dotfill \pageref{supple_sec:additional_results}\\[2pt]
\noindent\hspace*{1.5em}F.1\quad Additional Qualitative Results \dotfill \pageref{supple_sec:additional_results:qual}\\[2pt]
\noindent\hspace*{1.5em}F.2\quad Additional Quantitative Results \dotfill \pageref{supple_sec:additional_results:quan}\\[2pt]
\noindent\hspace*{1.5em}F.3\quad Challenging-based Evaluation  \dotfill \pageref{supple_sec:additional_results:challenge_eval}\\[2pt]

\vspace{0.5em}
\noindent\textbf{G\quad Limitations and Discussion} \dotfill \pageref{supple_sec:limitations_discussion}\\[2pt]
\noindent\hspace*{1.5em}G.1\quad Limitations \dotfill \pageref{supple_sec:limitations_discussion:limitations}\\[2pt]
\noindent\hspace*{1.5em}G.2\quad Discussion and Future Work \dotfill \pageref{supple_sec:limitations_discussion:discussion}\\[2pt]

\vspace{0.5em}
\noindent\textbf{H\quad Additional Related Works} \dotfill \pageref{supple_sec:additional_rel}\\[2pt]

\newpage


\section{Challenges of Cross-view Object Correspondence} 
\label{supple_sec:challenges}
\begin{figure*}[t!]
    \centering
\includegraphics[width=1\linewidth]{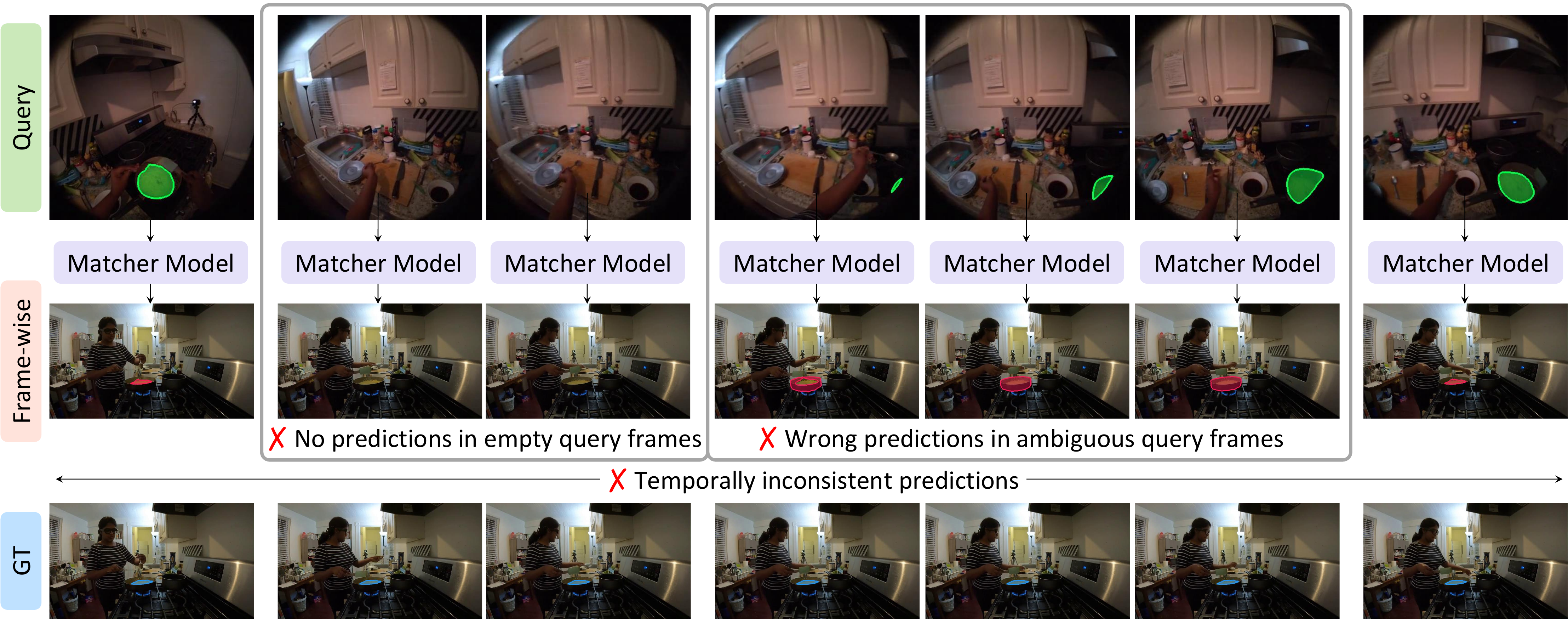}
    \vspace{-15pt}
    \caption{\textbf{Motivation for temporal reasoning in cross-view object correspondence.} Conventional frame-wise matching methods suffer from several failure modes, including empty predictions when the query object is absent and incorrect correspondences under ambiguous appearances. Since predictions are made independently for each frame, the resulting target correspondences often become temporally inconsistent.}
    \label{supple_fig:motivation}
\end{figure*}

\paragrapht{Motivation.}
We revisit the task of cross-view object correspondence~\cite{fu2025objectrelatorenablingcrossviewobject} and extend it to \textit{cross-view temporal object correspondence}. Prior to presenting our approach, we first discuss why this task is inherently challenging and motivate the need for temporal reasoning.  Fig.~\ref{supple_fig:motivation} illustrates these challenges in detail, complementing the paradigm comparison and examples in Fig.~\ref{fig:paradigm_comparison}. 

\paragrapht{Failure cases of the \textit{track-then-match} paradigm.} 
While tracking ensures temporal consistency on the query side, target predictions are produced independently per frame, with no explicit constraint enforcing temporal coherence across them. This leads to temporally inconsistent predictions on the target side. 
Therefore, we identify two primary failure modes on the query side in frame-level cross-view object correspondence. As shown in the second and third columns of Fig.~\ref{supple_fig:motivation}, when the query object is absent in a given query frame yet present in the target view, frame-level inference lacks a meaningful query signal, causing the model to either produce an empty prediction or hallucinate a correspondence to an incorrect object. The 4th to 6th columns further illustrate cases where the query object is visible but its appearance is highly ambiguous due to partial occlusion or an uninformative viewpoint, making target localization unreliable compared to cases where the object is fully observable. These failure modes compound to produce incomplete and inconsistent target predictions, ultimately resulting in temporally incoherent correspondences, as shown in the second row of Fig.~\ref{supple_fig:motivation}.

\paragrapht{From frame-level to temporal correspondence.} To address these limitations, we reformulate the problem as cross-view \textit{temporal} object correspondence. By incorporating temporal context beyond individual frames, our method alleviates the ambiguities inherent in frame-level inference and yields temporally consistent object correspondences across views. 

\section{\ours Details}
\label{supple_sec:impl_details}
\subsection{Architectural Details}
\label{supple_sec:impl_details:arch_details}
Our goal is to jointly model cross-view object correspondence and temporal propagation within a single framework. To this end, we build \ours on Wan2.1-T2V-1.3B~\cite{wan2025wan}, a large-scale pretrained video Diffusion Transformer (DiT). This backbone is chosen since it encodes rich priors on object motion, persistence, and appearance variations across viewpoints, which are essential for cross-view correspondence. The model consists of a 3D causal VAE, a text encoder~\cite{raffel2020exploring}, and a DiT. The VAE compresses input videos into latent tokens with spatial and temporal downsampling, and we keep both the VAE and text encoder frozen to preserve their pretrained representations, updating only the DiT. 

To adapt the pretrained model without destroying its learned priors, we employ LoRA adapters~\cite{hu2022lora} instead of full-finetuning. Specifically, we attach LoRA modules to all attention projections and feed-forward layers. This allows task-specific adaptation while keeping the original weights intact, which is critical for retaining the pretrained DiT's generative, spatiotemporal, and semantic priors. 
\subsubsection{Unified Ego-Exo Token Space}
To support cross-view reasoning, we represent paired clips $V^\text{q}=\{I_t^\text{q}\}_{t=0}^{T-1}$ and $V^{\bar{\text{q}}}=\{I_t^{\bar{\text{q}}}\}_{t=0}^{T-1}$ as a unified diptych input. 
Prior to VAE encoding, both ego and exo frames are resized to a common height $H$, ensuring that the diptych latent maintains a consistent spatial layout across views. Since ego cameras are typically fisheye with a near-square aspect ratio while exo cameras are standard with a wider aspect ratio, this height alignment results in $W^\text{q}\ne W^{\bar{\text{q}}}$ in general, where $W^\text{q}$ and $W^{\bar{\text{q}}}$ denote the width of each view, respectively. After VAE encoding, the latent grids of the two clips are concatenated along the width axis at every latent time step. 
Let $T_z$ denote the compressed temporal length after the VAE. 
If the query and target latents have size $T_z \times h \times w^\text{q}$ and $T_z \times h \times w^{\bar{\text{q}}}$, the resulting diptych latent has size $T_z \times h \times (w^\text{q}+w^{\bar{\text{q}}})$, where $h$ is the height and $w^\text{q}, w^{\bar{\text{q}}}$ are the widths of the corresponding latents.

The first-frame query mask $M_0^\text{q}$ is encoded in the same latent space and placed on the query-side region of the diptych. 
Since the query is provided only at the first frame, mask conditioning is injected at the corresponding latent frame, while all other temporal positions are zero-padded. 
The target-side region remains empty. 
This yields conditioning tokens that are spatially aligned with the query-view tokens and follow the same diptych layout.

All tokens are augmented with spatial and temporal positional embeddings. 
In particular, we assign RoPE~\cite{su2024roformer} based on each token’s frame-height-width coordinates on the diptych canvas. 
Since $z_\text{cond}$ exists only at the first frame, its tokens share the same RoPE coordinates as the first-frame RGB tokens, ensuring spatial and temporal alignment between conditioning and visual evidence.

This representation allows the DiT to process query-view tokens, target-view tokens, and conditioning tokens within a single spatiotemporal sequence. 
Standard full attention naturally induces both within-view and cross-view interactions, enabling information exchange across views at every layer without additional fusion modules.

\subsubsection{RGB-to-Mask Flow}
\label{supple_sec:impl_details:rgb2mask_details}
Within this unified token space, the model operates entirely in the latent domain. 
Given input videos $V^\text{q}$ and $V^{\bar{\text{q}}}$, the VAE encoder $\mathcal{E}$ produces RGB latents 
\[
z_\text{rgb} \in \mathbb{R}^{T_z\times h\times (w^\text{q} + w^{\bar{\text{q}}}) \times C},
\]
where $C$ denotes the channel dimension.

We construct a conditioning latent 
\[
z_\text{cond} \in \mathbb{R}^{1\times h\times (w^\text{q} + w^{\bar{\text{q}}})\times C}
\]
on the same diptych canvas:
\[
z_{\mathrm{cond}}[:, :, :w^\text{q}, :] = \mathcal{E}(M_0^\text{q}), 
\qquad
z_{\mathrm{cond}}[:, :, w^\text{q}:, :] = 0.
\]

The prediction target is the full diptych mask latent 
\[
z_\text{mask}\in\mathbb{R}^{T_z\times h \times (w^\text{q} + w^{\bar{\text{q}}}) \times C},
\]
which contains the mask sequences for both views over time.

The DiT takes $z_\text{rgb}$ and $z_\text{cond}$ as input and processes them as a single token sequence. 
We repurpose the DiT with a flow-based formulation by defining a transport from RGB latent space to mask latent space. 
Specifically, $z_\text{rgb}$ serves as the source state and $z_\text{mask}$ as the target state, and the model predicts the transport direction conditioned on $z_\text{cond}$. The predicted latent has the same diptych structure as $z_\text{rgb}$ and is decoded by the VAE decoder $\mathcal{D}$ to produce binary mask sequences in both views.

This formulation preserves the pretrained video DiT’s spatiotemporal and semantic priors while redirecting its generative capacity toward query-conditioned mask prediction. 
Because both views and all frames are processed jointly, the model can leverage cross-view evidence and temporal context, rather than relying on independent per-frame or per-view inference.

\subsection{Cross-View Object Grounding Details}
\label{supple_sec:impl_details:grounding_details}
We further incorporate three architectural components to improve grounding robustness.

\subsubsection{Focal Attention Alignment}
\begin{wrapfigure}[10]{r}{0.3\textwidth} 
    \vspace{-15pt}
    \centering
    \includegraphics[width=\linewidth]{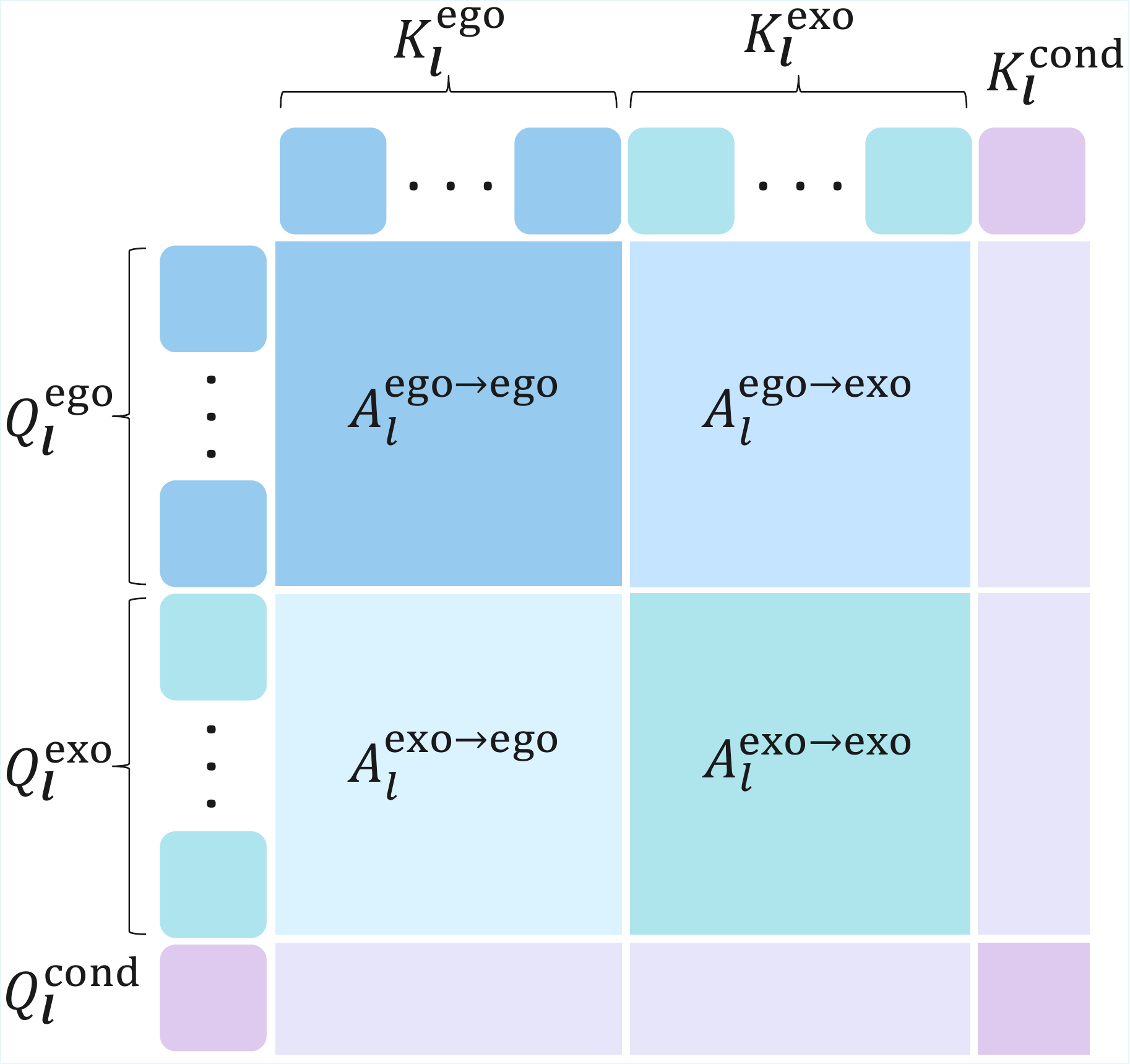}
    \vspace{-15pt}
    \caption{\textbf{3D full attention in video DiTs.}}
    \vspace{-15pt}
    \label{fig:3d_full_attention}    
\end{wrapfigure}

The unified token space provides cross-view attention blocks at every DiT layer, as shown in Fig.~\ref{fig:3d_full_attention}, but these blocks are not guaranteed to focus on the queried object. We therefore add an auxiliary loss that directly supervises target-to-query attention using foreground masks.

At layer $l$, let $A_l^{\bar{\text{q}} \rightarrow \text{q}} \in \mathbb{R}^{N_{\bar{\text{q}}} \times N_\text{q}}$ denote the cross-view attention block whose rows correspond to target-view query tokens and columns correspond to query-view key tokens. Here, $N_{\bar{\text{q}}}$ and $N_\text{q}$ are the numbers of target-view and query-view tokens. We downsample the query mask to the token resolution and obtain a soft foreground vector $w^\text{q} \in [0,1]^{N_\text{q}}$. We also downsample the target mask to define the set of target foreground tokens $\mathcal{F}^{\bar{\text{q}}}$.
We apply the same formulation to both Ego2Exo and Exo2Ego by swapping $q$ and $\bar{\text{q}}$. This loss does not introduce a separate matching module. It only biases the existing cross-view attention blocks to use the query-side object evidence more effectively.

\subsubsection{Hard Negative Conditioning}
To handle instance ambiguity, we introduce a negative conditioning signal. For a training chunk with another valid object instance in the same take, we encode that object's first-frame mask with the VAE and place it on the source side of the diptych as a negative condition. The model is then evaluated under both the positive query condition and the negative condition. The hard-negative loss penalizes the model when the prediction under the negative condition becomes too similar to the prediction under the positive condition.

This design explicitly exposes the model to competing same-scene objects. Same-category siblings can be filtered when their supervision is ambiguous, and if no valid sibling exists, the negative condition is sampled from a buffer. This makes the negative path a structural conditioning branch rather than only a loss-level modification.

\subsubsection{Local Context Crop}
The full diptych latent is spatially compressed by the VAE, so small objects may be represented by only a few tokens. To provide higher-resolution evidence, we construct an object-centered crop from the source view. Specifically, we compute the union bounding box of the query mask, dilate it by a margin ratio, resize the crop to a fixed resolution, and encode it with the same VAE used for the global input. The resulting crop latents are appended to the conditioning tokens rather than replacing the global diptych tokens. This allows the model to access both global scene context and local object details.

To preserve spatial consistency, the crop tokens are not treated as an independent image. We assign RoPE coordinates according to the crop location in the original full frame. Thus, a crop token retains its corresponding frame, height, and width position relative to the global diptych canvas. During training, the local context crop is used only for small-object chunks and is randomly dropped with a fixed probability, preventing the model from relying exclusively on local context crops.

\subsection{Additional Training Strategy}
\label{supple_sec:impl_details:training}
\subsubsection{Conditioning Mask Augmentation.}
We also augment the input query mask before VAE encoding. At each training step, the conditioning mask is sampled from three prompt types. The original mask is used as the strongest prompt, a downsampled and re-upsampled coarse mask simulates noisy annotation, and the bounding box mask provides weaker spatial supervision. All variants are encoded by the same VAE and inserted into the same query-side conditioning region of the diptych. This keeps the input interface unchanged while exposing the model to different prompt qualities.

\subsection{Implementation Details}
\label{supple_sec:impl_details:impl}
We use Wan2.1-T2V-1.3B~\cite{wan2025wan} as the backbone. The diptych latent has spatial resolution $h \times (w^\text{q} + w^{\bar{\text{q}}})$ and temporal length $T_z$, determined by VAE compression. Each training sample contains $T=29$ frames, which are encoded into $T_z=8$ latent frames, with ego scene at $448\times448$ and exo scene at $448\times800$.

We train using AdamW~\cite{loshchilov2017decoupled} optimizer using a learning rate of $1\mathrm{e}{-5}$ and weight decay of $0.01$. The effective batch size is $16$ and training is performed on $4$ H100 GPUs with mixed precision.

The loss weights are set to $\lambda_{\text{FM}}=1.0$, $\lambda_{\text{align}}=0.5$, $\lambda_{\text{neg}}=0.5$, and $\lambda_{\text{cycle}}=1.0$. Foreground tokens are defined by thresholding masks at $0.5$, and focal weighting uses $\gamma=2$.

For local context crops, we enlarge the query bounding box by a margin ratio of $1.2$ and resize the crop to $224 \times 224$ before VAE encoding. At inference, we use $50$ diffusion steps with classifier-free guidance scale $3.5$.

\section{Pseudo-Label Curation Pipeline Details}
\label{supple_sec:curation_details}
\subsection{Bidirectional Propagation Details}
\label{supple_sec:curation_details:propagation_details}
Extending sparse annotations to dense mask tracks via a propagation model is a standard practice in video segmentation benchmarks such as VISOR, UVO, MUVOD, and HD-EPIC~\cite{darkhalil2022epickitchensvisorbenchmarkvideo, wang2021unidentifiedvideoobjectsbenchmark, wei2025muvodnovelmultiviewvideo, perrett2025hdepichighlydetailedegocentricvideo}. Following this established protocol, we adopt SAM2-hiera-large~\cite{ravi2024sam}, a state-of-the-art propagation model, as the backbone of our pseudo-label generation pipeline.
Given two consecutive annotated anchor frames, we perform bidirectional propagation: forward propagation from the earlier anchor and backward propagation from the later anchor. For each temporal chunk bounded by anchor frames, we initialize SAM2 with the ground-truth mask and propagate masks across all intermediate frames, producing forward and backward candidate mask sequences.
However, propagation errors frequently occur in ego-view videos due to occlusion, hand-object interaction, and rapid viewpoint changes. This motivates the need for a subsequent reliability estimation and refinement process.

\subsection{Unreliable Chunk Selection Details}
\label{supple_sec:curation_details:chunk_selection_details}
To identify unreliable regions, we exploit the disagreement between forward and backward propagations. Intuitively, if propagation is stable, both directions should produce consistent masks for the same frame.
A straightforward way to estimate reliability is to reconstruct a mask by composing forward and backward propagations and compare it against the anchor ground truth, as done in prior work~\cite{darkhalil2022epickitchensvisorbenchmarkvideo}. However, this approach evaluates consistency only at the anchor endpoints, and therefore does not directly reflect the reliability of individual intermediate frames. In particular, errors such as drift that are corrected near the anchor remain undetected under this formulation.
Instead, we directly compare the forward and backward masks at each frame. For each frame within a chunk, we compute the mIoU between the forward-propagated mask and the backward-propagated mask. We then aggregate this frame-wise consistency into a chunk-level reliability score by averaging the mIoU values across the chunk.
Chunks whose average mIoU falls below a predefined threshold are marked as unreliable and selected for further refinement. In addition, to better localize failure regions, we identify consecutive frames with low mIoU and treat them as unreliable temporal segments within the chunk.
This bidirectional consistency check serves as a lightweight yet effective proxy for propagation reliability. Unlike endpoint-based reconstruction metrics, our approach directly measures disagreement at each frame and is therefore more sensitive to local failures.

\subsection{Cross-view Geometric Verification Details}
\label{supple_sec:curation_details:cross-view_details}

For chunks identified as unreliable, we resolve the ambiguity between forward and backward propagation candidates using cross-view geometric verification.
At each timestamp, we obtain a reference mask from the exo view by independently applying SAM2 propagation. Since exo cameras are static and exhibit significantly less motion blur and occlusion compared to ego views, their propagated masks are empirically more stable and serve as a reliable cross-view reference.
Instead of explicit geometric triangulation, which is not well-suited for large viewpoint disparities and dynamic scenes, we leverage VGGT~\cite{wang2025vggt} features to establish correspondence across views. Specifically, we extract dense feature maps from a pretrained VGGT encoder for both forward and backward ego-view candidates and the exo-view reference.
For each candidate, we compute cosine similarity between the feature vectors within the masked regions. Concretely, we first perform mask-aware average pooling over the feature maps to obtain a representative feature vector for each region, and then compute cosine similarity between the ego-view candidate and the exo-view reference. In addition, we compute a dense similarity map and evaluate its spatial consistency to ensure that the correspondence is not driven by spurious local matches.
The final matching score is obtained by combining the region-level similarity and the spatial consistency of the similarity map. We then select the forward or backward candidate with the higher matching score as the refined mask for that frame.
This verification step is applied only when the object is visible in the exo view. When the object is fully occluded or absent in the exo view, this step is skipped, and the pipeline falls back to within-view semantic verification.

\subsection{Within-view Semantic Verification Details}
\label{supple_sec:curation_details:within-view_details}

The cross-view verification stage provides a relative selection between candidates but does not guarantee absolute correctness. Moreover, it cannot be applied when exo-view references are unavailable. To address these limitations, we introduce a within-view identity verification stage based on semantic consistency.
For each selected candidate mask, we compare it against the nearest ego-view anchor ground-truth mask. In practice, we consider both preceding and succeeding anchors and select the one that yields higher semantic similarity.
We extract DINOv3~\cite{simeoni2025dinov3} features from both the candidate region and the anchor ground-truth region. Similar to the geometric verification stage, we apply mask-aware average pooling to obtain region-level feature vectors, and compute cosine similarity between them.
If the similarity exceeds a predefined threshold, the candidate is accepted as valid pseudo ground truth. Otherwise, it is discarded. This step effectively filters out cases where the selected mask corresponds to a different object instance or exhibits severe semantic drift.
When cross-view verification is skipped, due to exo-view occlusion, we directly apply within-view verification to both forward and backward candidates. Among the candidates that satisfy the similarity threshold, we select the one with higher similarity. If neither candidate passes the threshold, the entire segment is discarded.

\subsection{Dataset Examples and Statistics}
\label{supple_sec:curation_details:dataset_examples}

\begin{figure*}[t!]
    \centering
\includegraphics[width=1\linewidth]{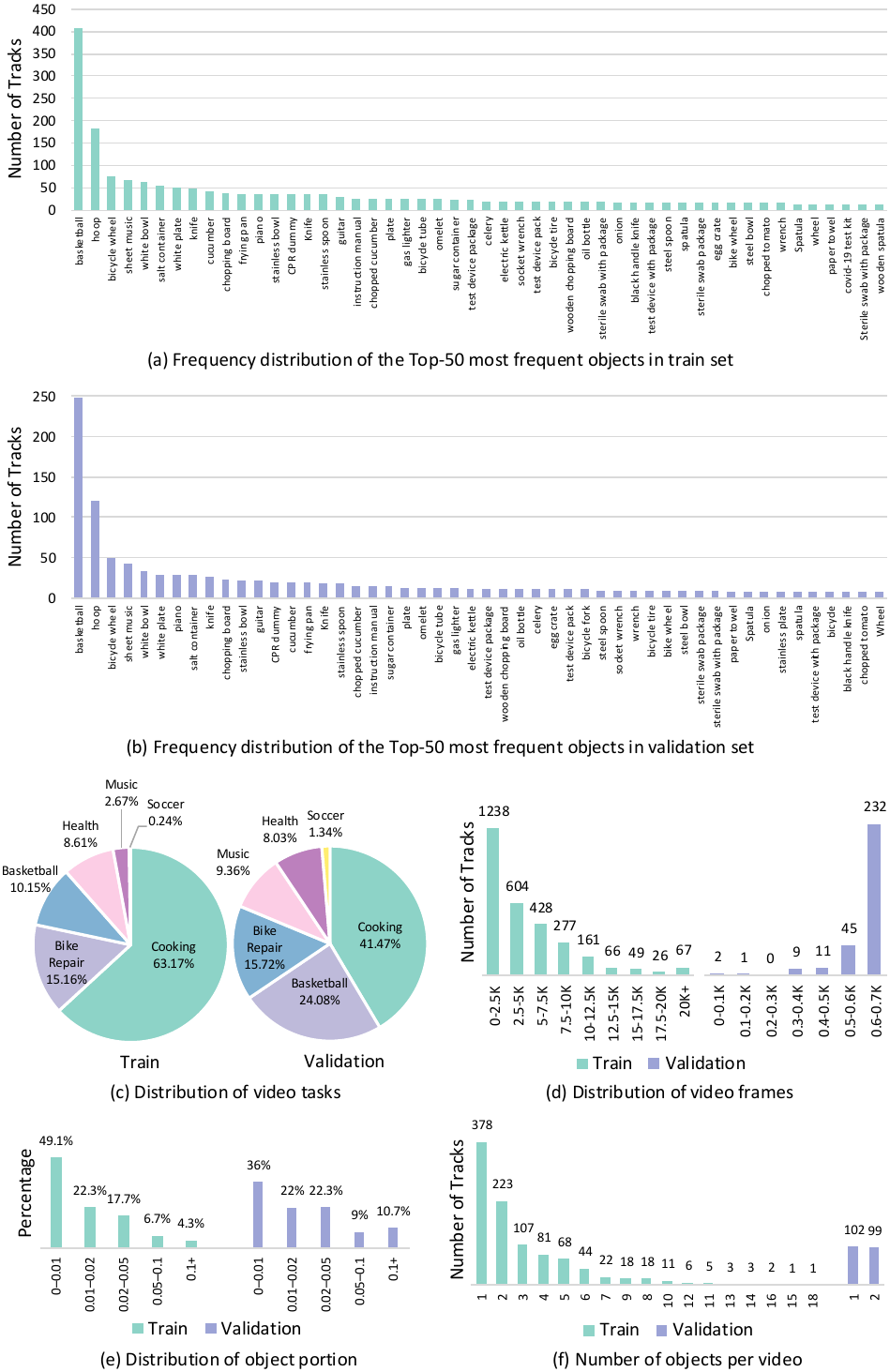}
    \vspace{-15pt}
    \caption{\textbf{Dataset statistics.}}
    \label{supple_fig:dataset_statistics}
\end{figure*}

\begin{figure*}[t!]
    \centering
\includegraphics[width=1\linewidth]{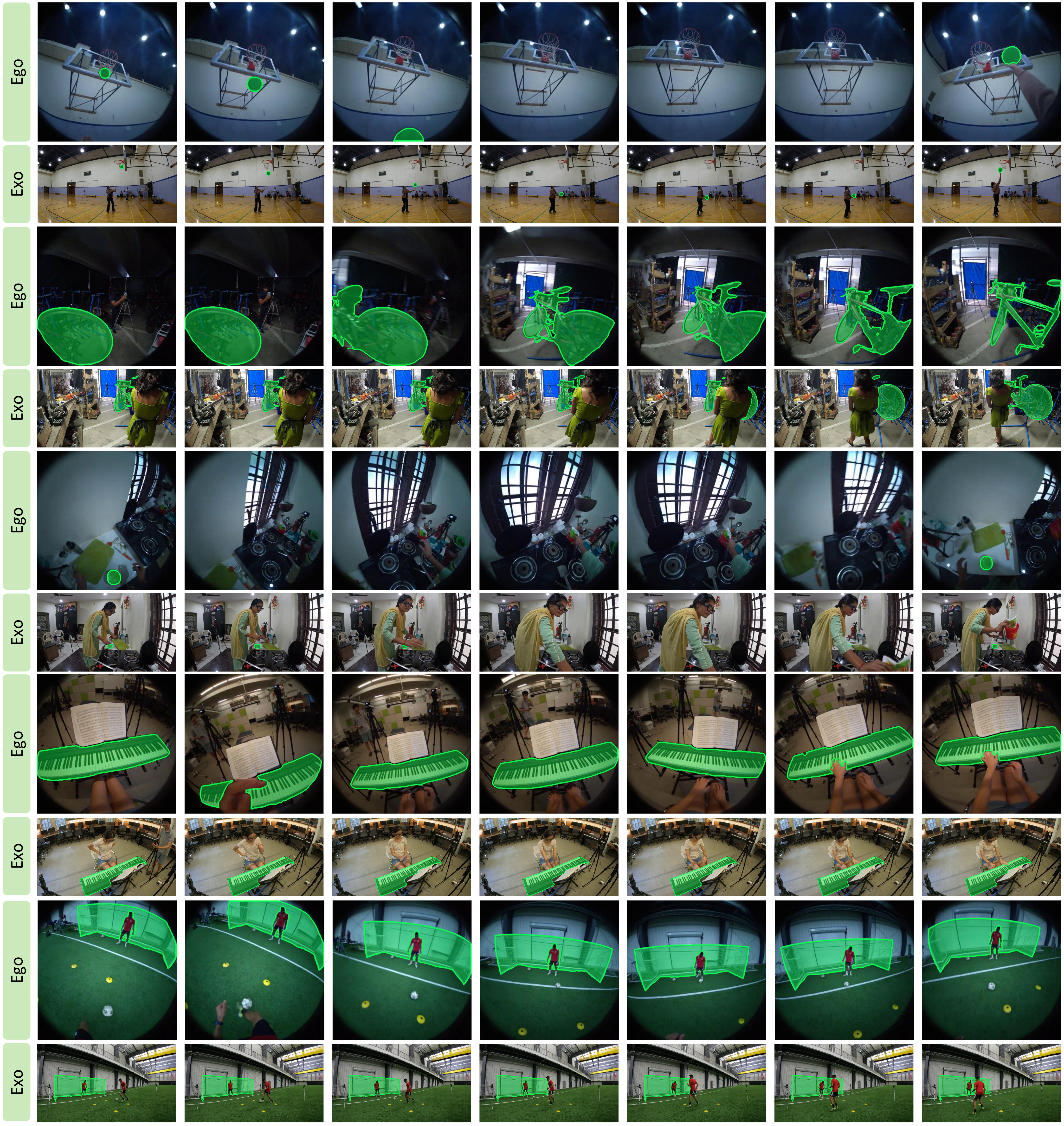}
    \vspace{-15pt}
    \caption{\textbf{Example sample of curated dataset.} Example samples from our curated cross-view temporal correspondence dataset across several tasks. Each pair of rows corresponds to a single sample, showing synchronized ego and exo views with temporally dense correspondence annotations.}
    \label{supple_fig:dataset_example}
\end{figure*}

\begin{figure*}[t!]
    \centering
    \vspace{-15pt}
\includegraphics[width=1\linewidth]{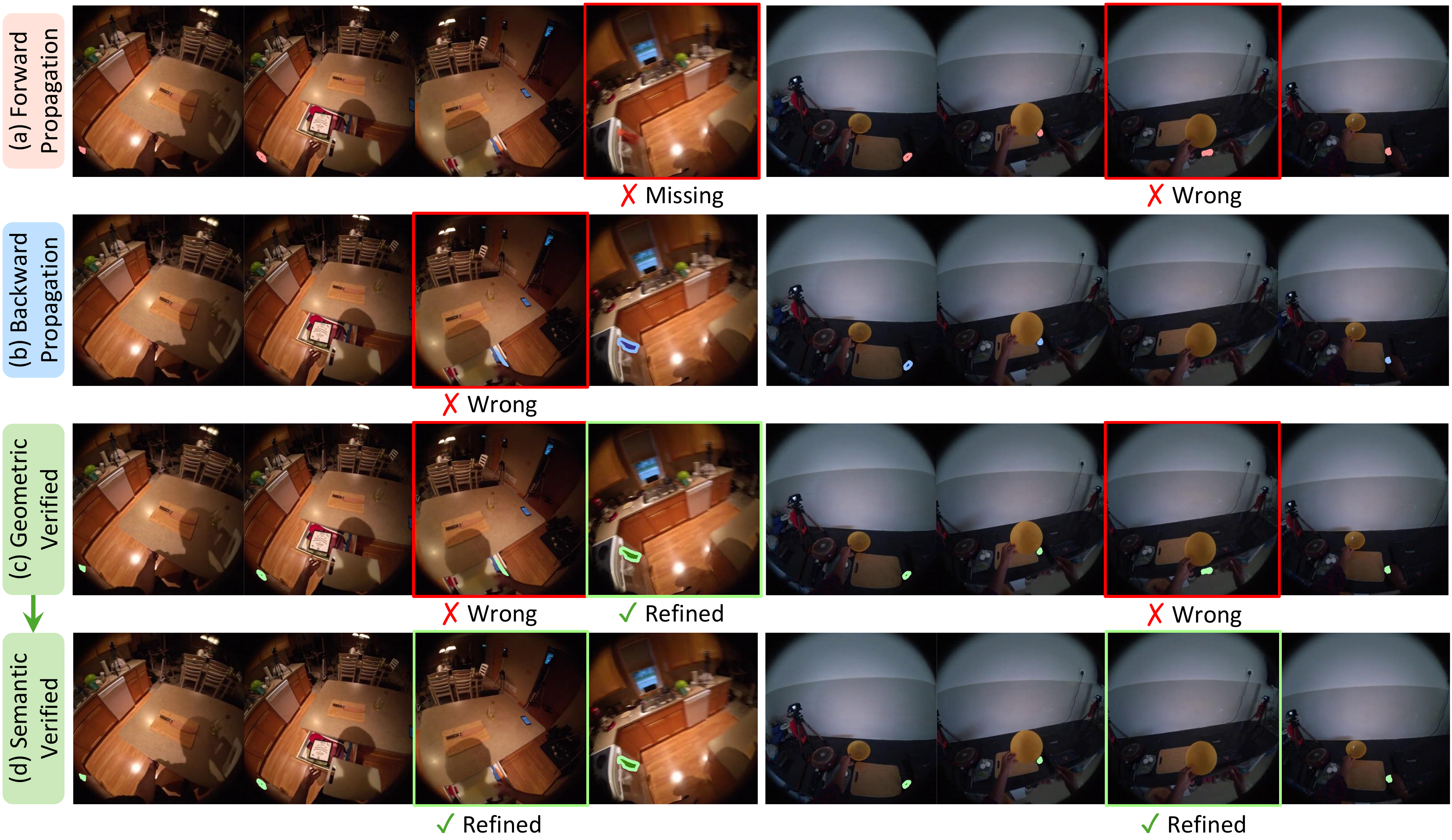}
    \vspace{-15pt}
    \caption{\textbf{Qualitative results of dataset curation pipeline.} (a) Forward-only propagation produces temporally inconsistent results, including missing masks and inaccurate object localization. (b) Backward propagation still exhibits similar failure cases, including incorrect propagations and ambiguous predictions. (c) Our cross-view geometric verification stage selects geometrically consistent correspondences from bidirectional propagation results. (d) Our within-view semantic verification stage further refines the selected masks by removing semantically incorrect predictions that cannot be resolved through geometry alone. These results demonstrate that both cross-view geometric verification and within-view semantic verification are necessary for constructing reliable dense correspondence annotations.}
    \label{supple_fig:curation_example}
\end{figure*}

Applying the proposed pseudo-label curation pipeline results in a large-scale pseudo-dense dataset comprising 708 training videos and 191 validation videos, with a total of 14M and 172K annotated frames, respectively. Here, a video corresponds to a unique take, and each track is defined as a unique (take, object) pair. As a result, a single video may contain multiple tracks when multiple objects are annotated. The total number of tracks is 3K in the training set and 300 in the validation set, and the frame count is computed by aggregating all frames across tracks. Compared to the original sparse annotations, our pipeline significantly increases temporal coverage, enabling frame-level supervision across diverse scenarios. As shown in Fig.~\ref{supple_fig:dataset_statistics}, the dataset covers a wide range of activities, including cooking, sports, and daily behaviors. Objects typically occupy a small portion of the frame, making segmentation and cross-view correspondence more challenging under large viewpoint changes and occlusions. When constructing the validation split, we carefully sample videos to preserve distributions of task categories (Fig.~\ref{supple_fig:dataset_statistics} (c)) and object scale (Fig.~\ref{supple_fig:dataset_statistics} (e)) similar to those of the training set, ensuring a consistent yet diverse evaluation protocol. For the validation set, human verification was conducted through track-level screening rather than exhaustive frame-by-frame correction. Tracks exhibiting identity switches, severe drift, or target loss were regenerated, and 10\% of frames from accepted tracks were additionally sampled for manual inspection to iteratively refine the annotation pipeline. Overall, the resulting dataset is both large-scale and structurally complex, making it well-suited for evaluating robust ego-exo cross-view video understanding. Fig.~\ref{supple_fig:dataset_example} provides the examples from the final curated dataset, and Fig.~\ref{supple_fig:curation_example} provides qualitative results of dataset curation pipeline.

\section{Evaluation}
\label{supple_sec:evaluation}

\subsection{Evaluation Dataset}
\label{supple_sec:evaluation:evaluation_dataset}
Our validation set consists of 191 takes and 300 tracks, yielding 172K densely annotated frames per direction, 344K in total across Ego2Exo and Exo2Ego. All validation tracks underwent track-level human screening rather than exhaustive frame-by-frame correction, as detailed in Appendix~\ref{supple_sec:curation_details:dataset_examples}. The official Ego-Exo4D~\cite{grauman2024ego} Correspondence benchmark provides 1 FPS annotations over 202 validation takes and 291 test takes, containing approximately 290K and 420K masks across both views, respectively. Recent methods construct their evaluation pools from these splits. For instance, V$^2$-SAM~\cite{pan2025v} evaluates on the test split with approximately 200K (Ego2Exo) and 219K (Exo2Ego) masks, while O-MaMa~\cite{mur2025mama} adopts the same official benchmark protocol as V$^2$-SAM on the test v2 split. ObjectRelator~\cite{fu2025objectrelatorenablingcrossviewobject} evaluates on the validation split, using 31K (Ego2Exo) and 36K (Exo2Ego) annotated frames. In terms of scale, our 344K masks are comparable in scale to prior evaluation setups. 
Moreover, we preserve the underlying data distribution through sampling strategies that maintain task category proportions (Fig.~\ref{supple_fig:dataset_statistics}(c)) and object scale distributions (Fig.~\ref{supple_fig:dataset_statistics}(e)), ensuring consistency with existing benchmarks.
Furthermore, Ego-Exo4D takes are on average approximately three minutes long and often contain repetitive segments. Evaluating every frame would therefore over-represent duplicate states without improving diversity, while also incurring unnecessary computational cost. To address this, we construct each validation track with up to 600 frames: shorter tracks are kept in full, while longer tracks are uniformly subsampled.

\subsection{Evaluation Metric}
\label{supple_sec:evaluation:evaluation_metric}
Our task requires both accurate mask prediction and stable identity preservation across views and time. Therefore, as we mentioned in Sec.~\ref{sec:experimental_setup} of the main paper, we evaluate performance in terms of per-frame segmentation quality and temporal consistency. 

\paragrapht{Per-frame Segmentation Quality.}
For per-frame segmentation quality, we use region similarity $\mathcal{J}$, contour accuracy $\mathcal{F}$ and their average $\mathcal{J\& F}$. $\mathcal{J}$ measures region overlap between the predicted mask and ground-truth mask. This evaluates whether the model segments the correct object region. $\mathcal{F}$ measures boundary alignment between predicted and ground-truth contours, which captures whether the mask shape is accurate. We report $\mathcal{J\& F} = \frac{1}{2}(\mathcal{J} + \mathcal{F})$ as a balanced score for region and boundary quality. 

We also report Location Error (LE) and Visibility Accuracy (VA), following Ego-Exo4D~\cite{grauman2024ego}. LE measures the distance between predicted and ground-truth object centers. This is useful since a mask may have low overlap but still localize the correct object region, or conversely drift toward a wrong instance. VA measures whether the model correctly predicts object visibility. This is critical in ego-exo videos since objects can disappear due to occlusion, leave the field of view, or reappear later.

\paragrapht{Temporal Consistency.} 
We additionally evaluate temporal consistency using Recall@IoU. Per-frame metrics measure mask quality at each frame, but they do not capture whether the same object identity is preserved throughout a sequence. Given an IoU threshold $p$, Recall@IoU evaluates whether a sequence that starts with a correct prediction continues to maintain it. Specifically, we first select sequences that satisfy $\mathcal{J}_0^{(i)} \geq p$ at the first frame, and then measure whether the prediction remains above the threshold for all subsequent frames:
\begin{equation}
\mathrm{R}@p = \frac{1}{N_p} \sum_{i=1}^{N_p}
\mathbb{1}\left(\min_t \mathcal{J}_t^{(i)} \geq p\right),
\end{equation}
where $N_p$ is the number of sequences satisfying $\mathcal{J}_0^{(i)} \geq p$. 

This metric directly measures temporal retention conditioned on a correct initialization, rather than initial matching accuracy. In the main paper, we report $\mathrm{mR}@\mathrm{IoU}$, the mean of $\mathrm{R}@p$ over thresholds $\{0.1, 0.3, 0.5, 0.7\}$. If no sequence satisfies the initialization condition at a given threshold, the metric is undefined and reported as NaN.

\section{Analyses}
\label{supple_sec:analyses}
\subsection{Analysis for Design Choice}
\label{supple_sec:analyses:analysis_design_choice}
\begin{figure*}[t!]
    \centering
\includegraphics[width=1\linewidth]{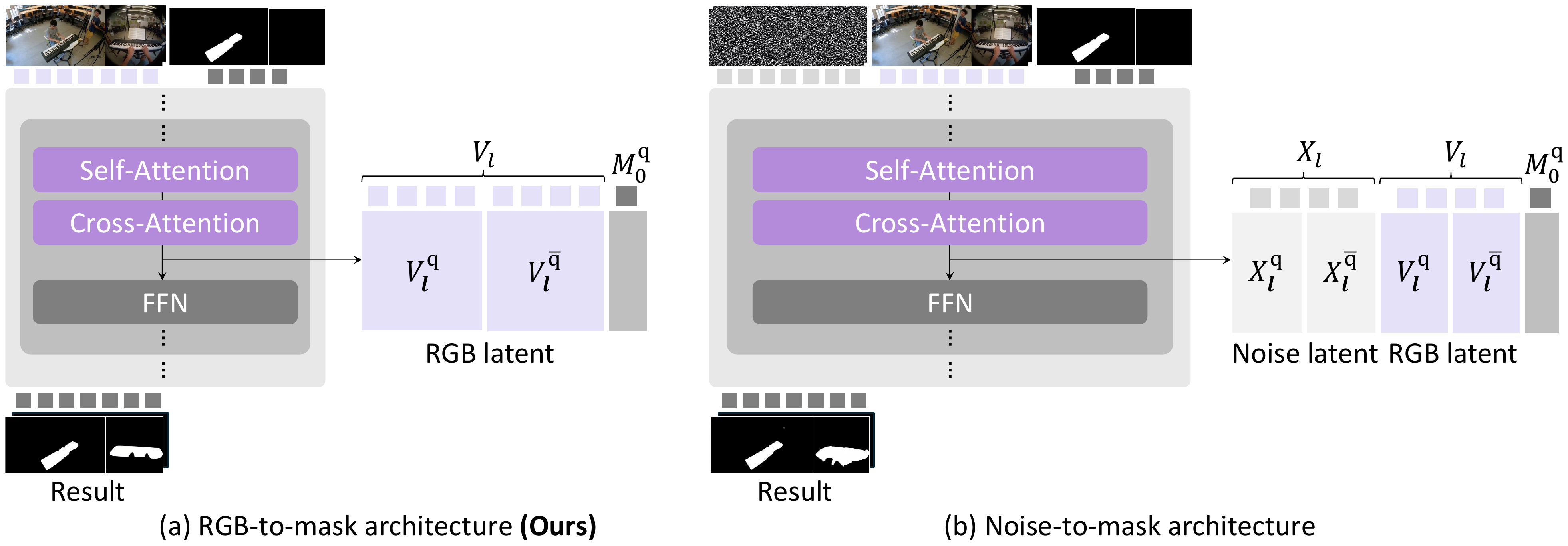}
    \vspace{-15pt}
    \caption{\textbf{Difference between noise-to-mask and RGB-to-mask design.} (a) Our RGB-to-mask architecture directly predicts mask flow from RGB latents, allowing the transport trajectory to focus entirely on mask prediction. (b) In contrast, the noise-to-mask design predicts masks by denoising from noise while conditioning on RGB latents, following the standard diffusion formulation. As a result, the model implicitly undergoes an intermediate noise-to-RGB transport before mask prediction. Implicit noise-to-RGB transition is further visualized through PCA analysis in Fig.~\ref{supple_fig:pca_visualization}.}
    \label{supple_fig:flow_matching_design_difference}
\end{figure*}

\begin{figure*}[t!]
    \centering
\includegraphics[width=1\linewidth]{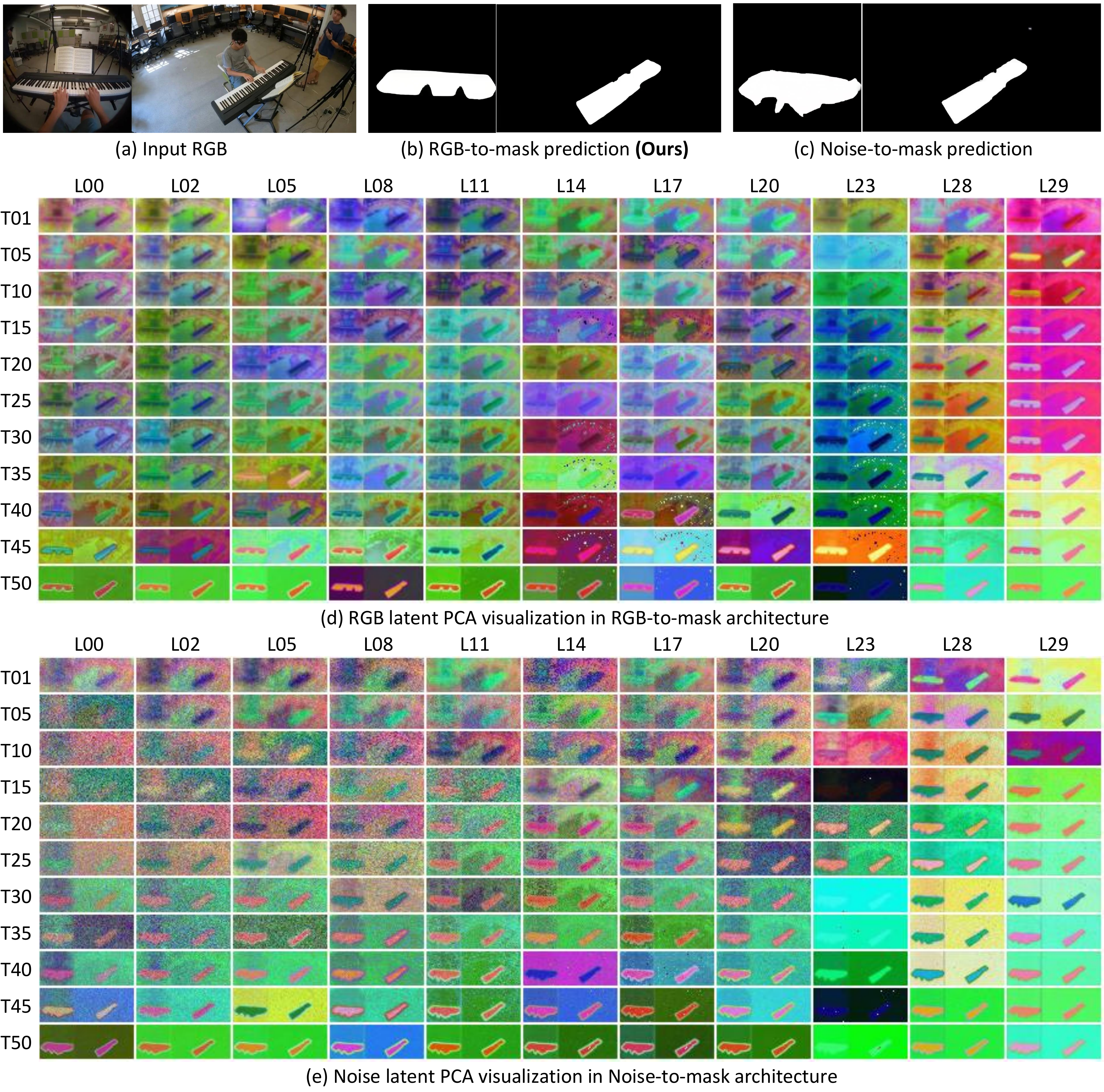}
    \vspace{-15pt}
    \caption{\textbf{Visualization of feature PCA.} Results are shown in the exo-query, ego-target setting. (a) shows the input RGB frames, (b) shows the final prediction from the RGB-to-mask formulation, and (c) shows the final prediction from the noise-to-mask formulation. (d) visualizes the RGB latent trajectory \(V_l\) from the RGB-to-mask architecture in Fig.~\ref{supple_fig:flow_matching_design_difference}(a), while (e) visualizes the noise latent trajectory \(X_l\) from the noise-to-mask architecture in Fig.~\ref{supple_fig:flow_matching_design_difference}(b). (e) exhibits an implicit transition through RGB-like representations in the early timesteps before converging toward mask prediction. Columns denote transformer layers and rows denote denoising timesteps.}    
    \label{supple_fig:pca_visualization}
\end{figure*}

We also ablate the choice of initial state in the flow matching formulation. As described in Sec.~\ref{sec:rgb2mask} in main paper and Sec.~\ref{supple_sec:impl_details:rgb2mask_details} in Appendix, \ours adopts an \textit{RGB-to-mask} formulation that sets the RGB latent $z_\text{rgb}$ as the source state of the flow. A natural alternative is \textit{Noise-to-mask} prediction, where the source state is replaced with random Gaussian noise $z_0 \sim \mathcal{N}(0, I)$, and the model predicts the transport field from noise directly to the target mask latent, conditioned with RGB latents.

Fig.~\ref{supple_fig:flow_matching_design_difference} illustrates the conceptual difference between the two formulations. In noise-to-mask (Fig.~\ref{supple_fig:flow_matching_design_difference} (b)), the model must implicitly reconstruct the RGB scene structure from the conditioning signal before it can perform mask prediction. This intermediate step introduces an implicit noise-to-RGB transport within the trajectory, which is not directly supervised and can produce noisy intermediate representations.


To analyze this behavior, we visualize the per-layer, per-timestep PCA features of the intermediate latent states for both formulations. Specifically, Fig.~\ref{supple_fig:pca_visualization} (d) visualizes the RGB latent trajectory $V_l$ from the RGB-to-mask design in Fig.~\ref{supple_fig:flow_matching_design_difference} (a), while Fig.~\ref{supple_fig:pca_visualization} (e) visualizes the noise latent trajectory $X_l$ from the noise-to-mask design in Fig.~\ref{supple_fig:flow_matching_design_difference} (b). As shown in Fig.~\ref{supple_fig:pca_visualization} (d), the RGB-to-mask trajectory shows significantly cleaner and more structured intermediate representations from the earliest timesteps. Crucially, semantic object regions are captured more precisely in the early layers and timesteps, indicating that grounding in the RGB latent allows the model to focus on cross-view object correspondence from the outset rather than recovering scene structure implicitly. In contrast, the Noise-to-mask trajectory (Fig.~\ref{supple_fig:pca_visualization} (e)) exhibits an implicit transition through RGB-like representations in the early timesteps before converging toward mask prediction. This suggests that the model first spends capacity reconstructing scene content from noise before attending to the target object.

This difference in intermediate representations is also reflected in the final predictions. As shown in Fig.~\ref{supple_fig:pca_visualization} (b), the RGB-to-mask formulation produces masks that better preserve fine object details and boundaries, closely aligning with the input RGB observation in Fig.~\ref{supple_fig:pca_visualization} (a). In contrast, the Noise-to-mask prediction in Fig.~\ref{supple_fig:pca_visualization} (c) tends to be less precise and exhibit degraded structure, indicating weaker semantic grounding. These results confirm that the RGB-to-mask formulation provides a more effective initialization for cross-view temporal mask prediction. By starting from a source state that already encodes scene layout and object appearance, the transport field is entirely devoted to converting visual evidence into query-conditioned masks, leading to more reliable and semantically grounded predictions.

Beyond representational quality, the two formulations also differ substantially in computational cost. In the Noise-to-mask formulation, the model requires both the noise latent and the RGB latent as separate inputs, as the noise latent serves as the initial state of the flow, while the RGB observation must be provided as an additional conditioning signal. 
In the diptych setting with paired ego-exo videos, the Noise-to-mask formulation processes both noise and RGB tokens from the two views, along with mask-conditioning tokens.
Let $N$ denote the total number of RGB tokens across both views, and $N_{\mathrm{cond}}$ denote the number of mask-conditioning tokens, including the query-side mask and target-side zero-mask regions.
Excluding optional local-crop tokens, the Noise-to-mask formulation requires approximately $2N + N_{\mathrm{cond}}$ tokens.
In contrast, the RGB-to-mask formulation directly sets the RGB latent as the source state, eliminating the need for separate noise tokens and reducing the sequence length to approximately $N + N_{\mathrm{cond}}$.
Since self-attention cost scales quadratically with sequence length, the relative attention cost is proportional to $\left(\frac{2N + N_{\mathrm{cond}}}{N + N_{\mathrm{cond}}}\right)^2$.
Thus, the Noise-to-mask formulation incurs higher attention cost, approaching four times that of RGB-to-mask when $N$ dominates $N_{\mathrm{cond}}$.

As a result, the RGB-to-mask formulation offers advantages in both representation quality and computational efficiency in our setting.
It provides a more semantically grounded and structured initialization that enables effective cross-view object correspondence from the earliest timesteps, while reducing the token sequence length and substantially lowering attention cost.


\subsection{Analysis for Attention Alignment}
\label{supple_sec:analyses:attention}
\begin{figure*}[t!]
    \centering
\includegraphics[width=1\linewidth]{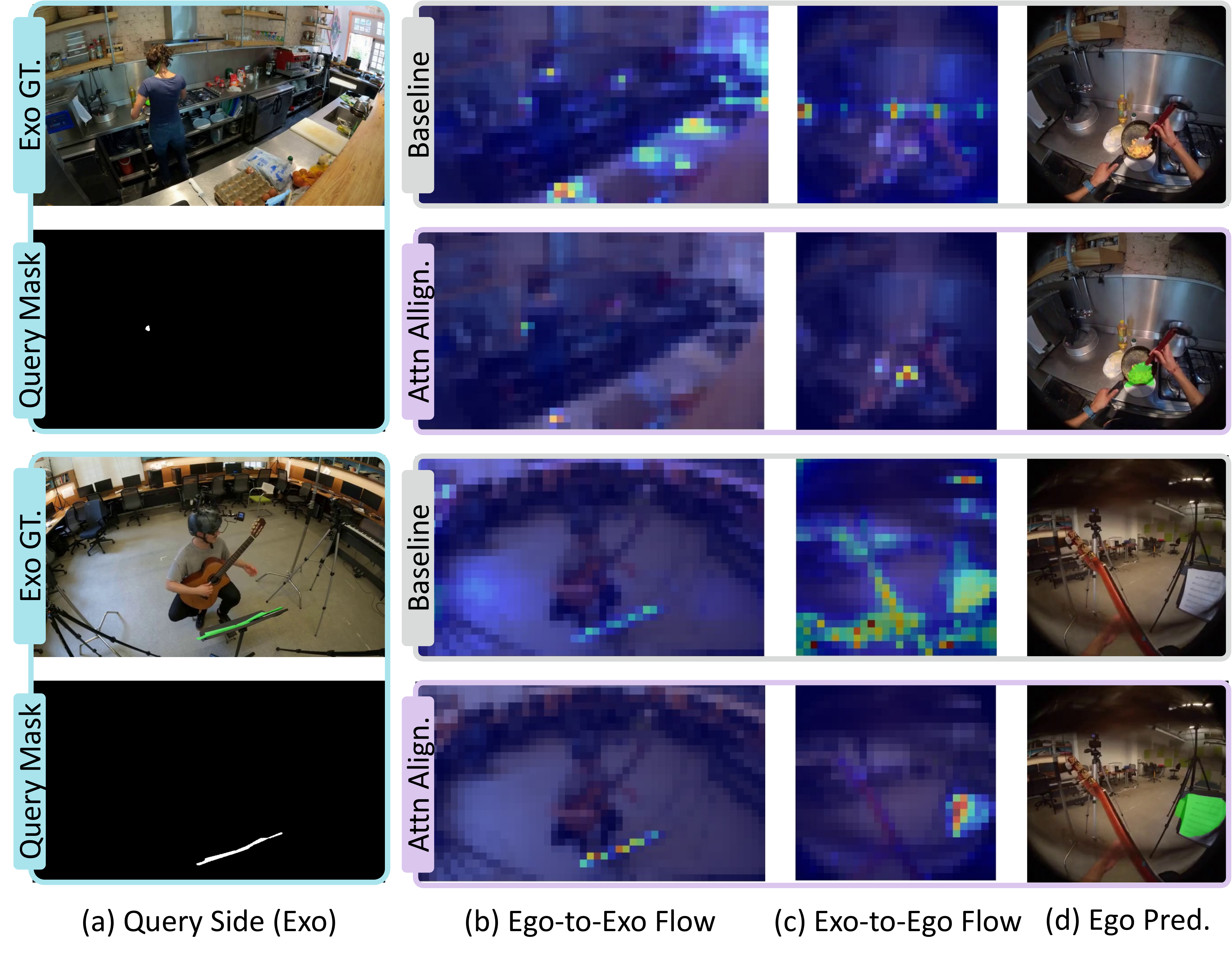}
    \vspace{-15pt}
    \caption{\textbf{Visualization of cross-view attention.} 
    (a) shows the exocentric query frame and query mask used as input. 
    (b) visualizes the cross-view attention map \(A^{\text{ego}\rightarrow \text{exo}}\), indicating which exocentric regions the ego-side target attends to when establishing correspondence. 
    (c) visualizes the reverse attention map \(A^{\text{exo}\rightarrow \text{ego}}\). (d) shows the resulting ego-side prediction mask. The first and third rows correspond to the baseline without attention alignment, while the second and fourth rows show the results after applying focal attention alignment. Without alignment, attention is broadly dispersed across irrelevant regions, resulting in inaccurate or missing predictions. In contrast, focal attention alignment produces compact and localized attention patterns that better align with the target object, leading to more accurate cross-view correspondence predictions.}    \label{supple_fig:attention_visualization}
\end{figure*}

As we mentioned in Sec.~\ref{sec:cross-view-obj-grounding} of the main paper, we propose focal attention alignment as one of our cross-view object grounding strategies to facilitate more reliable object correspondence within the unified token space. Here, we provide a detailed motivation and analysis of this design choice, along with analysis through attention visualization. 

\subsubsection{Unified Token Space and the Implicit Matching Problem}
In our framework, query and target tokens are jointly encoded as $z_\text{rgb}$ in a unified token space via diptych, where all tokens are processed within a single shared representation, regardless of their view or temporal signal. The 3D full attention mechanism enables unrestricted interaction across views and timesteps, allowing each token to attend to any other token in the sequence, as illustrated in Fig.~\ref{fig:3d_full_attention}. This architectural choice is precisely what enables cross-view and cross-temporal reasoning, as discussed in the main paper. However, while the model is given the capacity to establish correspondences across views and time, the architecture itself imposes no explicit constraint that target object tokens must attend to the semantically correct query object tokens. In other words, correspondence is left entirely to implicit matching via self-attention which globally operates on the entire sequence, and there is no supervisory signal that directly encourages the model to focus on the correct object region during the attention computation. 

\subsubsection{Attention Dispersion as a Root Cause of Failure}
We hypothesize that this reliance on purely implicit matching becomes a critical bottleneck, especially in complex scenes. When multiple objects are present, or when the target scene contains distractors that share visual similarity or semantic similarity with the query object, attention tends to disperse broadly across semantically irrelevant regions rather than concentrating on the correct correspondence. This dispersion prevents the model from extracting a sufficiently discriminative cross-view representation, resulting in incorrect or incomplete mask predictions. Critically, this failure mode is not resolved by simply scaling model capacity or training data, since without an explicit grounding signal, the model has no way to resolve ambiguous attention patterns. 

Fig.~\ref{supple_fig:attention_visualization} directly validates this hypothesis via cross-view attention visualization. Column (a) shows the query-side input, depicting the exocentric view and the query mask used as input. Columns (b) and (c) visualize the cross-view attention maps $A^{\text{ego}\rightarrow \text{exo}}$ and $A^{\text{exo}\rightarrow \text{ego}}$ from Fig.~\ref{fig:3d_full_attention}. Note that these maps are computed by restricting the query positions to tokens within the ground-truth object mask region and averaging the resulting attention weights across those query positions. Concretely, (b) captures where and how strongly the ego-side target object region attends to the exo view when establishing cross-view correspondence, that is, which exocentric regions the model relies upon when attempting to localize the object on the ego side. Column (c), conversely, reflects how consistently the exo-side query object region attends back to the ego view, representing the degree to which the query provides coherent and localized evidence for the correspondence. Column (d) shows the resulting ego-side prediction mask, allowing a direct comparison between the attention distribution and the final output. 

The first and third rows of Fig.~\ref{supple_fig:attention_visualization} correspond to the baseline without attention alignment, and the second and fourth rows show the results after applying focal attention alignment. In the baseline, both $A^{\text{ego}\rightarrow \text{exo}}$ and $A^{\text{exo}\rightarrow\text{ego}}$ exhibit broad, diffuse attention patterns that fail to concentrate on the correct object region, spreading across background clutter and semantically unrelated areas. The predictions in column (d) directly reflect this dispersion, yielding masks that are either mislocalized, fragmented, or entirely absent. In contrast, after applying focal attention alignment, the attention in both directions becomes sharply concentrated on the semantically correct regions, forming compact and well-localized attention patterns that closely conform to the ground-truth object boundary. Surprisingly, the spatial structure of resulting prediction in column (d) closely mirrors the attended region, providing direct visual evidence of a causal relationship between attention localization and prediction quality.

\subsubsection{Why Focal Attention Alignment Works}
These observations motivate the design of focal attention alignment as an explicit supervision signal that encourages cross-view attention $A^{\text{ego}\rightarrow \text{exo}}$ and $A^{\text{exo}\rightarrow\text{ego}}$ to concentrate on the correct object region during training. Rather than relying solely on the downstream mask prediction loss to implicitly shape attention, focal attention alignment directly regularizes the attention distribution toward the ground-truth object region, providing a more direct and localized training signal. This not only improves attention localization but also yields more discriminative cross-view representations, as the model is guided to extract object-centric features rather than scene-level statistics. The result is a measurable improvement in correspondence accuracy, especially in complex scenes, where implicit matching is most prone to failure, as reflected in the quantitative results in the main paper. 

\subsection{Effect of Local Context Crops on Small Objects}
\label{supple_sec:analyses:small_object_analysis}
\begin{wraptable}{r}{0.35\columnwidth}
\vspace{-8pt}
\centering
\caption{\textbf{Small-object analysis.}
J\&F on the bottom 25\% smallest objects.}
\label{tab:small_object}
\vspace{3pt}
\scriptsize
\setlength{\tabcolsep}{5pt}
\renewcommand{\arraystretch}{1.15}

\begin{tabular}{lcc}
\hline
Method & Ego2Exo & Exo2Ego \\
\hline
w/o Crop & 0.211 & 0.223 \\
\rowcolor{mypurple}
\textbf{w/ Crop (Ours)}
& \textbf{0.376}
& \textbf{0.327} \\
\hline
\end{tabular}

\vspace{-6pt}
\end{wraptable}
To further analyze the effectiveness of local context crops, we evaluate J\&F on the smallest-object subset, defined as the bottom 25\% by object size.
As shown in Tab.~\ref{tab:small_object}, local context crops substantially improve mask quality in both directions,
with a particularly large relative gain in Ego2Exo.
This trend is consistent with the greater ambiguity of
egocentric queries, where small objects are often partially
occluded, truncated, or affected by rapid camera motion.
By providing higher-resolution object-centric evidence
alongside global scene context, local crops help mitigate
the loss of fine-grained information caused by latent-space
compression.
Nevertheless, the remaining difficulty in recovering precise
object boundaries suggests that local context crops alleviate,
rather than fully resolve, the limitations of compressed
latent representations.

\section{Additional Results}
\label{supple_sec:additional_results}
\subsection{Additional Qualitative Results}
\label{supple_sec:additional_results:qual}
\begin{figure*}[t!]
    \centering
\includegraphics[width=1\linewidth]{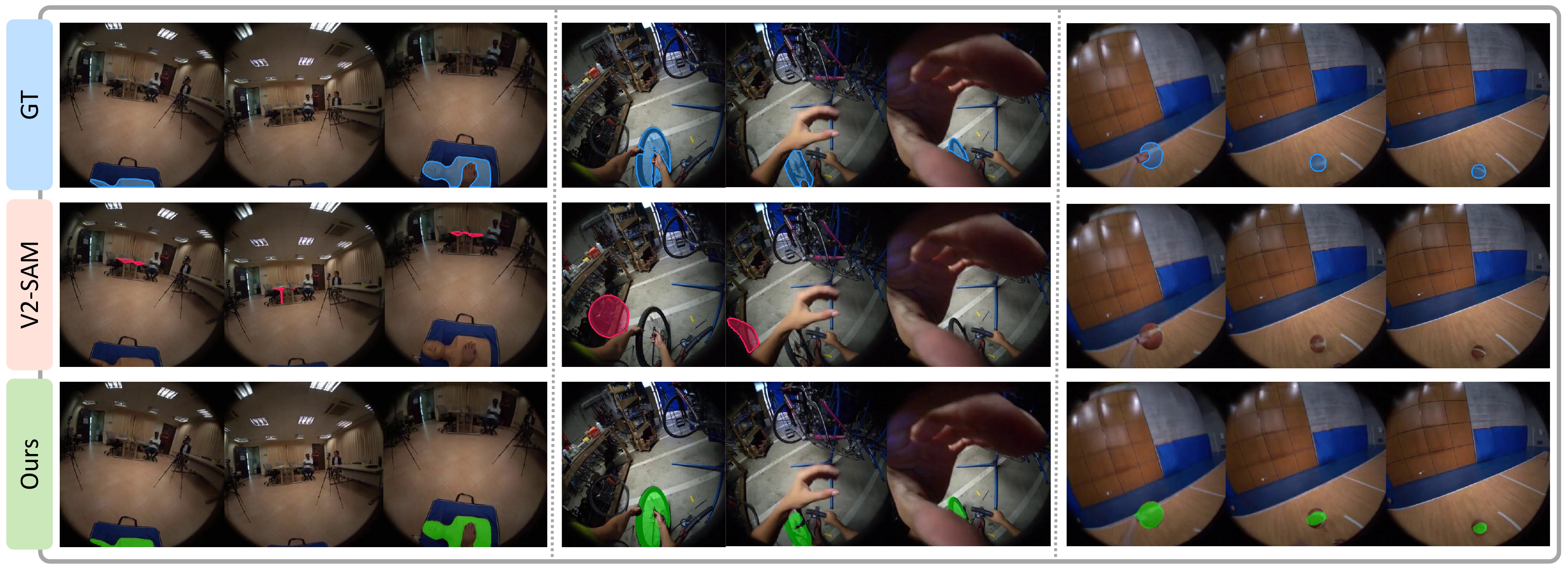}
    \vspace{-15pt}
\caption{\textbf{Additional qualitative results of comparison with V$^2$-SAM.}
Results are shown in the exo-query, ego-target setting. As a match-then-track pipeline, V$^2$-SAM heavily depends on the initial cross-view match, making it difficult to refine incorrect or incomplete matches during tracking. In contrast, our method more reliably localizes the target object's presence across frames, producing temporally stable masks under challenging cross-view appearance changes.}
    \label{supple_fig:v2sam_comparison}
\end{figure*}

We provide additional qualitative comparisons with V$^2$-SAM in Fig.~\ref{supple_fig:v2sam_comparison}. 
Although V$^2$-SAM achieves strong scores on mask quality related metrics such as $\mathcal{J}$, $\mathcal{F}$, and $\mathcal{J}\&\mathcal{F}$, its match-then-track pipeline is highly dependent on the quality of the initial cross-view match. When the initial localization is incorrect or incomplete, the subsequent tracking stage has limited ability to recover or refine the target object, causing the error to persist over time. In contrast, our method directly reasons over cross-view temporal correspondence and more reliably localizes the target object's presence across frames. This leads to more temporally consistent predictions, even though the generated masks may occasionally have less precise object boundaries due to the compressed latent representation of the video DiT backbone.
\begin{figure*}[t!]
    \centering
\includegraphics[width=1\linewidth]{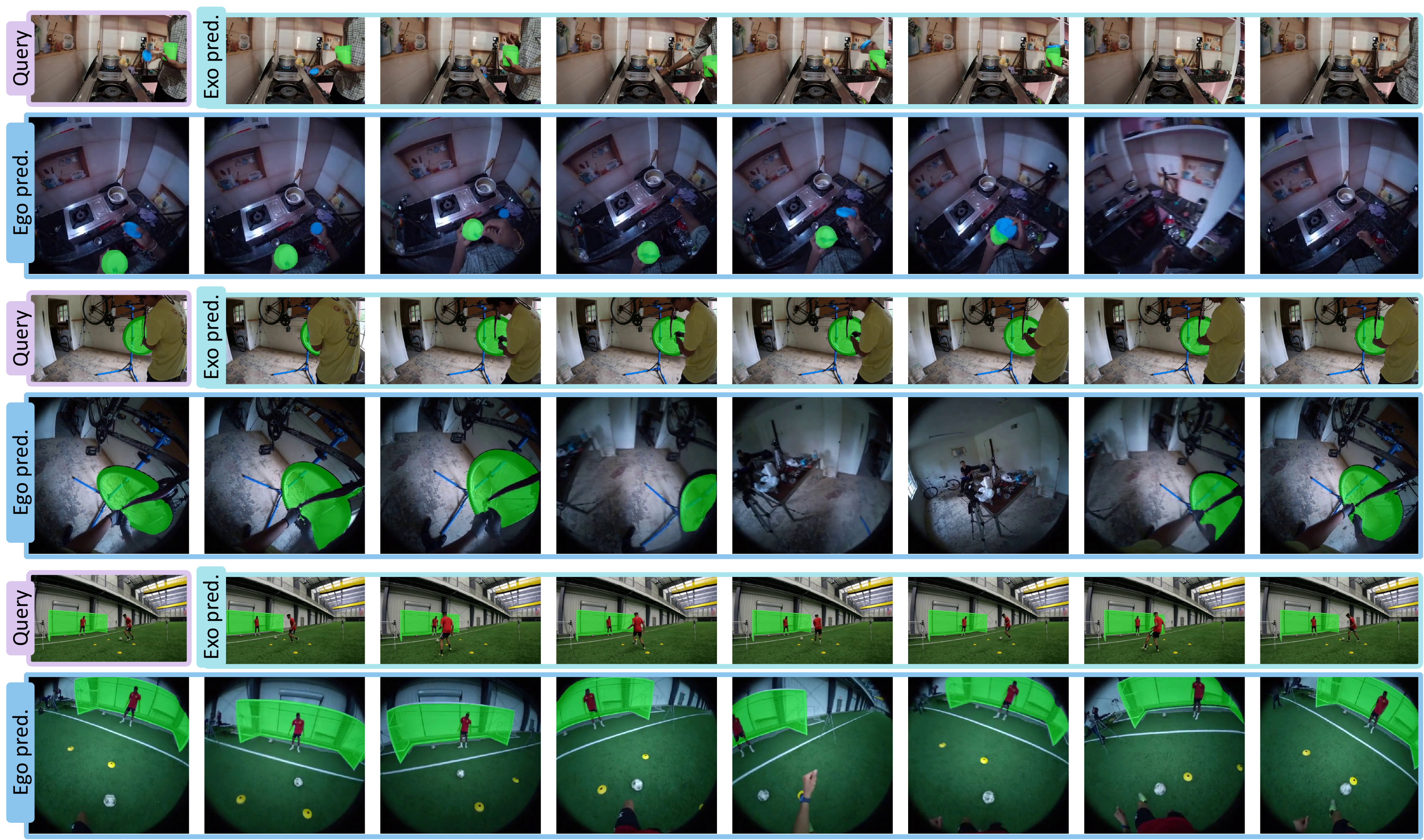}
    \vspace{-15pt}
    \caption{\textbf{Additional qualitative results of Exo2Ego.} Results are shown in the exo-query, ego-target setting. Each pair of rows corresponds to a single sample, where the first row shows the query exocentric frames and propagated exo predictions, and the second row shows the corresponding ego-side predictions over time.}
    \label{supple_fig:additional_qual}
\end{figure*}

\begin{figure*}[t!]
    \centering
\includegraphics[width=1\linewidth]{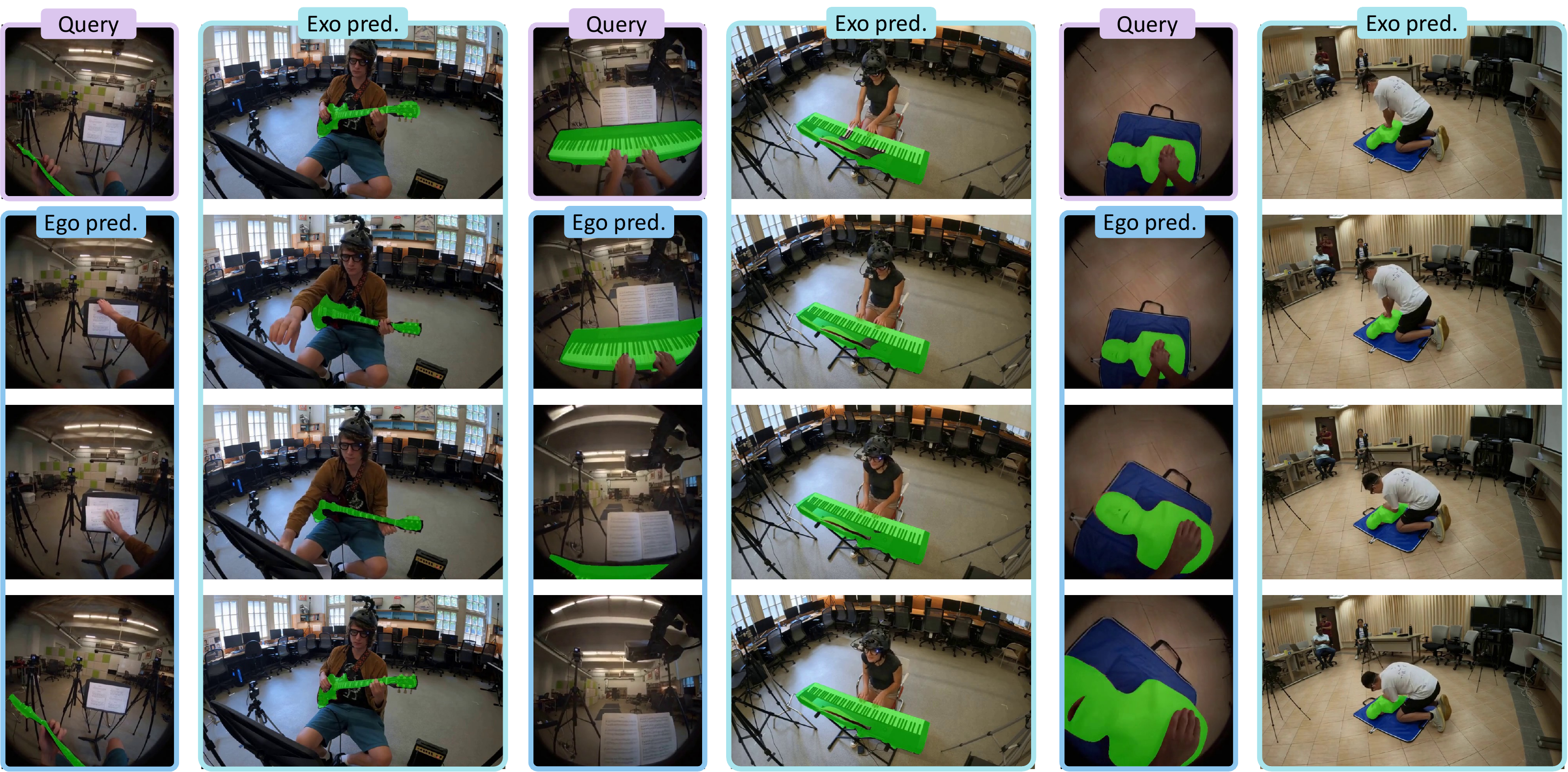}
    \vspace{-15pt}
    \caption{\textbf{Additional qualitative results of Ego2Exo.} Results are shown in the ego-query, exo-target setting. Each pair of columns corresponds to a single sample, where the left column shows the ego query frames and ego-side predictions, and the right column shows the corresponding exo-side predictions over time.}
    \label{supple_fig:additional_qual_ego2exo}
\end{figure*}

Fig.~\ref{supple_fig:additional_qual} and Fig.~\ref{supple_fig:additional_qual_ego2exo} provide additional qualitative comparisons supplementing the results presented in the main paper. Fig.~\ref{supple_fig:additional_qual} illustrates Exo2Ego correspondences, where exocentric query frames generally offer complete and unambiguous object observations, and the primary challenge lies in localizing the corresponding object in the dynamic and cluttered egocentric target view. Fig.~\ref{supple_fig:additional_qual_ego2exo} presents Ego2Exo correspondences, which pose a greater challenge as the egocentric query frames frequently suffer from object truncation, occlusion, and rapid viewpoint changes. As evidenced by the qualitative results in Fig.~\ref{supple_fig:motivation}, prior methods struggle to produce temporally consistent target predictions under such conditions, often failing entirely when the query object is absent or ambiguous in individual frames. In contrast, our method yields coherent and accurate correspondences across both directions, demonstrating the effectiveness of temporal reasoning in resolving the ambiguities inherent to each setting.

\subsection{Additional Quantitative Results}
\label{supple_sec:additional_results:quan}

\begin{table*}
\centering
\caption{
\textbf{Quantitative comparison on EgoExo4D (Recall@p).}
Track-then-Match (T$\rightarrow$M) and Match-then-Track (M$\rightarrow$T) are reported for prior methods.
Our method is independent of pipeline choice. $\text{N/A}$ indicates missing checkpoints. $\text{NaN}$ indicates that no samples satisfy the IoU threshold $p$ at the first frame.
}
\resizebox{1\linewidth}{!}{
\begin{tabular}{l c | cccc | cccc}
\toprule
\multirow{2}{*}{Method} & \multirow{2}{*}{Pipeline}
& \multicolumn{4}{c|}{\textbf{Ego2Exo}}
& \multicolumn{4}{c}{\textbf{Exo2Ego}} \\
&
& R@0.1 & R@0.3 & R@0.5 & R@0.7
& R@0.1 & R@0.3 & R@0.5 & R@0.7 \\
\midrule

\multicolumn{10}{c}{\textit{Track-then-Match (T$\rightarrow$M)}} \\
\midrule
PSALM             & T$\rightarrow$M & 0.637 & 0.657 & 0.671 & 0.568 & 0.546 & 0.505 & 0.488 & 0.466 \\
XSegTx            & T$\rightarrow$M & 0.707 & 0.694 & \textbf{0.999} & NaN     & 0.721 & 0.703 & 0.677 & 0.607 \\
XView-XMem        & T$\rightarrow$M & 0.727 & 0.697 & 0.730 & 0.614 & 0.815 & 0.795 & 0.755 & 0.729 \\
XView-XMem+XSegTx & T$\rightarrow$M & 0.851 & 0.814 & 0.780 & 0.752 & 0.766 & 0.754 & 0.759 & 0.748 \\
ObjectRelator     & T$\rightarrow$M & 0.822 & 0.799 & 0.792 & 0.770 & 0.707 & 0.700 & 0.691 & 0.684 \\
O-MaMa            & T$\rightarrow$M & N/A    & N/A    & N/A    & N/A    & 0.606 & 0.586 & 0.548 & 0.483 \\
V$^2$-SAM         & T$\rightarrow$M & 0.734 & 0.708 & 0.689 & 0.653 & 0.700 & 0.681 & 0.673 & 0.677 \\
\midrule

\multicolumn{10}{c}{\textit{Match-then-Track (M$\rightarrow$T)}} \\
\midrule
PSALM             & M$\rightarrow$T & 0.967 & 0.933 & 0.943 & 0.911 & 0.856 & 0.861 & 0.832 & 0.805 \\
XSegTx            & M$\rightarrow$T & 0.871 & 0.825 & 0.671 & 0.524 & 0.904 & 0.897 & 0.884 & 0.851 \\
XView-XMem        & M$\rightarrow$T & 0.928 & 0.884 & 0.925 & 0.899 & 0.908 & 0.899 & 0.876 & 0.863 \\
XView-XMem+XSegTx & M$\rightarrow$T & 0.970 & 0.953 & 0.942 & 0.938 & 0.918 & 0.908 & 0.899 & \underline{0.877} \\
ObjectRelator     & M$\rightarrow$T & \underline{0.978} & \textbf{0.975} & 0.969 & \textbf{0.953} & \underline{0.919} & 0.907 & \underline{0.904} & \textbf{0.884} \\
O-MaMa            & M$\rightarrow$T & N/A    & N/A    & N/A    & N/A    & 0.905 & 0.884 & 0.850 & 0.805 \\
V$^2$-SAM         & M$\rightarrow$T & 0.968 & 0.966 & 0.963 & 0.921 & 0.918 & \underline{0.915} & 0.895 & 0.862 \\
\midrule
\cellcolor{mypurple} \textbf{Ours} & \cellcolor{mypurple} \textbf{Unified} & \cellcolor{mypurple}\textbf{0.979} & \cellcolor{mypurple}{0.959} & \cellcolor{mypurple}{0.949} & \cellcolor{mypurple}{0.927}  & \cellcolor{mypurple}\textbf{0.949}  & \cellcolor{mypurple}\textbf{0.932}  & \cellcolor{mypurple}\textbf{0.907}  & \cellcolor{mypurple}{0.855} \\
\bottomrule
\end{tabular}
}
\label{tab:recall_unified}

\end{table*}

Tab.~\ref{tab:recall_unified} presents a per-threshold breakdown of $\text{Recall@IoU}$ on EgoExo4D~\cite{grauman2024ego} for both $\text{Ego2Exo}$ and $\text{Exo2Ego}$ directions, complementing the mean $\text{Recall@IoU}$ $(\text{mR@IoU})$ reported in the Tab.~\ref{tab:main_quan_unified} of the main paper. 

The performance gap is particularly pronounced in the Ego2Exo direction, which we attribute to the inherently higher ambiguity of egocentric query frames. Egocentric views are characterized by frequent object truncation, rapid camera motion, and heavily occluded or partially visible objects, making it substantially harder to extract reliable query representations on a per-frame basis. This stands in contrast to the Exo2Ego direction, where exocentric query frames tend to provide more stable and complete object observations, partially mitigating the limitations of frame-level inference. 

These characteristics directly manifest in the failure modes of both existing paradigms, \textit{track-then-match} and \textit{match-then-track} pipelines. In the $\text{T}\rightarrow\text{M}$ paradigm, although tracking yields temporally consistent query object tracks, the per-frame matching to the target view is performed independently at each timestep with no mechanism to enforce temporal coherence on the target side. The degraded query signal in egocentric query frames thus directly corrupts each frame-level match, and the absence of any cross-frame aggregation, that is temporal context aggregation, leaves these errors unrecoverable. In the $\text{M}\rightarrow\text{T}$ paradigm, the initial match quality is entirely contingent on the frame-level query signal, and the subsequent tracker~\cite{ravi2024sam} inevitably propagates any erroneous initialization throughout the sequence. 

In contrast, our unified framework addresses these limitations by reasoning over the full temporal extent of the query sequence rather than individual frames. By aggregating complementary object observations across time, our method is able to compensate for frames in which the query object is absent or ambiguous, constructing a richer and more discriminative query representation that remains robust even under the challenging viewpoint conditions of egocentric capture. Overall, XTrack achieves strong temporal retention across both directions, with particularly competitive performance under challenging egocentric query conditions. 

\subsection{Challenge-based Evaluation}
\label{supple_sec:additional_results:challenge_eval}
\begin{table*}[t]
\centering
\caption{
\textbf{Challenge-based evaluation.}
Performance under delayed target appearance and temporally
disjoint visibility. T$\rightarrow$M and M$\rightarrow$T
denote Track-then-Match and Match-then-Track, respectively.
N/A indicates unavailable results.
}
\vspace{5pt}
\label{tab:challenge_eval}
\resizebox{\textwidth}{!}{%
\begin{tabular}{ll|ccccc|ccccc}
\hline
& & \multicolumn{5}{c|}{Ego2Exo}
    & \multicolumn{5}{c}{Exo2Ego} \\
Method & Pipeline
& $\mathcal{J}$ $\uparrow$ & $\mathcal{F}$ $\uparrow$ & $\mathcal{J\&F}$ $\uparrow$ & LE $\downarrow$ & VA $\uparrow$
& $\mathcal{J}$ $\uparrow$ & $\mathcal{F}$ $\uparrow$ & $\mathcal{J\&F}$ $\uparrow$ & LE $\downarrow$ & VA $\uparrow$ \\
\hline

\multicolumn{12}{l}{\cellcolor{gray!15}\textbf{Delayed Target Appearance}} \\
\hline
LM-EEC & M$\rightarrow$T
& 0.086 & 0.134 & 0.110 & 0.557 & 0.490
& 0.075 & 0.086 & 0.080 & 0.664 & 0.441 \\
ObjectRelator & M$\rightarrow$T
& 0.058 & 0.104 & 0.081 & 0.570 & 0.469
& 0.065 & 0.075 & 0.070 & 0.660 & 0.441 \\
V$^2$-SAM & M$\rightarrow$T
& 0.083 & 0.112 & 0.098 & 0.567 & 0.481
& 0.076 & 0.083 & 0.079 & 0.673 & 0.437 \\
O-MaMa & M$\rightarrow$T
& N/A & N/A & N/A & N/A & N/A
& 0.038 & 0.047 & 0.043 & 0.715 & 0.412 \\
\hline
LM-EEC & T$\rightarrow$M
& 0.111 & 0.207 & 0.159 & 0.551 & 0.486
& \underline{0.261} & \underline{0.298} & \underline{0.279} & 0.517 & \underline{0.540} \\
ObjectRelator & T$\rightarrow$M
& 0.102 & 0.193 & 0.148 & 0.550 & 0.477
& 0.244 & 0.284 & 0.264 & \underline{0.513} & 0.536 \\
V$^2$-SAM & T$\rightarrow$M
& \underline{0.198} & \underline{0.265} & \underline{0.231} & \underline{0.519} & \underline{0.508}
& 0.226 & 0.259 & 0.242 & 0.529 & 0.538 \\
O-MaMa & T$\rightarrow$M
& N/A & N/A & N/A & N/A & N/A
& 0.188 & 0.224 & 0.206 & 0.540 & 0.535 \\
\hline
\cellcolor{mypurple}\textbf{XTrack (Ours)}
& \cellcolor{mypurple}Unified
& \cellcolor{mypurple}\textbf{0.254}
& \cellcolor{mypurple}\textbf{0.299}
& \cellcolor{mypurple}\textbf{0.276}
& \cellcolor{mypurple}\textbf{0.138}
& \cellcolor{mypurple}\textbf{0.532}
& \cellcolor{mypurple}\textbf{0.368}
& \cellcolor{mypurple}\textbf{0.390}
& \cellcolor{mypurple}\textbf{0.379}
& \cellcolor{mypurple}\textbf{0.147}
& \cellcolor{mypurple}\textbf{0.594} \\
\hline

\multicolumn{12}{l}{\cellcolor{gray!15}\textbf{Temporally Disjoint Visibility}} \\
\hline
LM-EEC & M$\rightarrow$T
& 0.137 & 0.202 & 0.170 & 0.470 & 0.575
& 0.099 & 0.113 & 0.106 & 0.557 & 0.547 \\
ObjectRelator & M$\rightarrow$T
& 0.101 & 0.141 & 0.121 & 0.466 & 0.581
& 0.100 & 0.115 & 0.108 & 0.530 & 0.579 \\
V$^2$-SAM & M$\rightarrow$T
& 0.159 & 0.198 & 0.178 & 0.470 & 0.579
& 0.110 & 0.121 & 0.116 & 0.562 & 0.564 \\
O-MaMa & M$\rightarrow$T
& N/A & N/A & N/A & N/A & N/A
& 0.042 & 0.055 & 0.048 & 0.596 & 0.531 \\
\hline
LM-EEC & T$\rightarrow$M
& 0.213 & \textbf{0.338} & 0.275 & \underline{0.442} & \underline{0.588}
& \underline{0.249} & \underline{0.284} & \underline{0.266} & \underline{0.446} & \underline{0.629} \\
ObjectRelator & T$\rightarrow$M
& 0.213 & 0.264 & 0.239 & 0.590 & 0.422
& 0.216 & 0.240 & 0.228 & 0.596 & 0.441 \\
V$^2$-SAM & T$\rightarrow$M
& \textbf{0.255} & 0.299 & \textbf{0.277} & 0.573 & 0.441
& 0.221 & 0.246 & 0.234 & 0.603 & 0.437 \\
O-MaMa & T$\rightarrow$M
& N/A & N/A & N/A & N/A & N/A
& 0.202 & 0.228 & 0.215 & 0.607 & 0.436 \\
\hline
\cellcolor{mypurple}\textbf{XTrack (Ours)}
& \cellcolor{mypurple}Unified
& \cellcolor{mypurple}\underline{0.214}
& \cellcolor{mypurple}\underline{0.314}
& \cellcolor{mypurple}\underline{0.264}
& \cellcolor{mypurple}\textbf{0.137}
& \cellcolor{mypurple}\textbf{0.621}
& \cellcolor{mypurple}\textbf{0.311}
& \cellcolor{mypurple}\textbf{0.354}
& \cellcolor{mypurple}\textbf{0.332}
& \cellcolor{mypurple}\textbf{0.136}
& \cellcolor{mypurple}\textbf{0.646} \\
\hline
\end{tabular}
}
\end{table*}
To further examine the effectiveness of joint view-time reasoning,
we evaluate XTrack under two challenging visibility scenarios:
(1) delayed target appearance, where the query object is visible but the target appears later, and (2) temporally disjoint visibility, where the query object disappears before the target becomes visible. Using ground-truth visibility, we construct four subsets of approximately 10K frames each, covering both Ego2Exo and Exo2Ego directions. These cases account for approximately 10\% of the validation set.

As shown in Tab.~\ref{tab:challenge_eval}, XTrack achieves higher J\&F and VA, and lower LE than the compared baselines across both
scenarios and directions. In M$\rightarrow$T pipelines, ambiguous
or missing initial correspondences can propagate through subsequent
tracking, while T$\rightarrow$M pipelines are vulnerable to
inconsistent frame-wise matching under asymmetric visibility.
In contrast, XTrack jointly reasons over paired video sequences,
allowing later observations and cross-view evidence to resolve
ambiguous correspondences. These results further support the
benefit of joint view-time reasoning under challenging visibility
conditions.

\section{Limitations and Discussion}
\label{supple_sec:limitations_discussion}
\subsection{Limitations}
\label{supple_sec:limitations_discussion:limitations}
\paragrapht{Extremely small objects.}
\ours can struggle when the queried object is extremely small. Local context crops alleviate this issue by providing relatively higher-resolution local object evidence, but the final prediction is still produced in the compressed latent space of the video DiT. Therefore, if the object occupies only a few latent tokens, fine-grained boundaries or small appearance cues can still be lost. This mainly affects very small or heavily compressed objects, while the local context branch improves robustness in most small-object cases.

\paragrapht{Limited context length.}
\ours inherits the finite context length of the pretrained video DiT. For long videos, we process sequences in chunks and merge the predictions afterward. This enables long-sequence inference in practice, but the model cannot attend to arbitrarily distant frames within a single forward pass. Extending the temporal context or adding memory mechanisms would further improve long-range consistency.

\paragrapht{Inference cost.}
As a diffusion-based model, \ours requires iterative denoising at inference. This makes inference slower than feed-forward segmentation or tracking models. However, this cost comes from repurposing a generative video backbone for joint cross-view temporal prediction, and can likely be reduced with faster samplers~\cite{Lu_2025}, distillation~\cite{yin2024improveddistributionmatchingdistillation, zhu2024acceleratingvideodiffusionmodels, frans2025stepdiffusionshortcutmodels}, or fewer denoising steps~\cite{luo2023latentconsistencymodelssynthesizing, liu2024instaflowstephighqualitydiffusionbased, garcia2025finetuningimageconditionaldiffusionmodels}.

\subsection{Discussion and Future Work}
\label{supple_sec:limitations_discussion:discussion}
\paragrapht{Improving small-object representation.}
The difficulty with extremely small objects mainly arises from the spatial compression of latent representations. A promising direction is to incorporate multi-scale latent processing~\cite{ho2021cascadeddiffusionmodelshigh, gu2024matryoshkadiffusionmodels, zhang2025show1marryingpixellatent, jin2025pyramidalflowmatchingefficient}, where high-resolution tokens are selectively preserved or fused with global tokens. Another approach is to adapt the VAE~\cite{chen2025deepcompressionautoencoderefficient, kouzelis2025eqvaeequivarianceregularizedlatent} to retain more fine-grained spatial details for object-centric regions, or to introduce hierarchical tokenization~\cite{razavi2019generatingdiversehighfidelityimages, zhou2025hitvideohierarchicaltokenizersenhancing} that allocates more capacity to small objects.

\paragrapht{Extending temporal context.}
The current chunk-based processing limits the ability to model very long temporal dependencies within a single forward pass. Future work could explore memory-augmented architectures or recurrent token mechanisms that allow information to persist across chunks. Another direction is to integrate retrieval-based or keyframe-based conditioning, enabling the model to access long-range temporal context without increasing computational cost significantly.

\paragrapht{Reducing inference cost.}
The diffusion-based formulation introduces higher computational cost due to iterative denoising. This can be addressed by adopting more efficient sampling strategies, such as fewer-step solvers or distillation into feed-forward models. Additionally, hybrid approaches that combine diffusion for global reasoning with lightweight refinement networks may provide a favorable trade-off between efficiency and performance.

\section{Additional Related Works}
\label{supple_sec:additional_rel}
\paragrapht{Dataset Curation Pipeline.}
Several recent video segmentation datasets follow a similar high-level recipe of sparsely annotating anchor frames and propagating masks to obtain dense supervision, but they differ substantially in how each stage is instantiated.
For initial mask generation, UVO~\cite{wang2021unidentifiedvideoobjectsbenchmark}, VISOR~\cite{darkhalil2022epickitchensvisorbenchmarkvideo}, and HD-EPIC~\cite{perrett2025hdepichighlydetailedegocentricvideo} rely on manual annotation of anchor frames. Matrix-11K~\cite{jin2026matrixmasktrackalignment} replaces manual annotation by automatically selecting an anchor frame using a VLM and generating the initial mask with SAM2. MUVOD~\cite{wei2025muvodnovelmultiviewvideo} initializes masks on sparsely sampled keyframes using SAM~\cite{kirillov2023segment} conditioned on manually drawn bounding boxes, followed by manual refinement. In contrast, our pipeline is anchored entirely on the existing Ego-Exo4D annotations and does not introduce any additional manual labeling.

For temporal propagation, UVO and VISOR adopt bidirectional STM~\cite{oh2019videoobjectsegmentationusing} from sparse anchors, with logits weighted by the temporal distance to the nearest annotated frame. HD-EPIC propagates hand masks bidirectionally with SAM2~\cite{ravi2024sam} by inserting manually annotated frames into its conditional memory. Matrix-11K propagates SAM2 forward from a single VLM-selected anchor frame. MUVOD employs XMem~\cite{cheng2022xmemlongtermvideoobject} for bidirectional temporal propagation from corrected keyframes to recover content occluded in the reference frame. We also perform bidirectional propagation using SAM2.

A key distinction lies in how propagation failures are handled. UVO and MUVOD do not define a per-frame automatic failure criterion and instead rely on human intervention. HD-EPIC adopts a heuristic based on abrupt changes in mask area between consecutive frames to identify noisy predictions, followed by local correction and re-propagation. Matrix-11K filters at the video level via VLM verification rather than at the frame or mask level, providing no signal about the reliability of individual propagated frames. VISOR is the only prior work that introduces an explicit quantitative criterion: each interpolated mask is validated by propagating it back to both annotated endpoints and computing the average $\mathcal{J}\&\mathcal{F}$, after which masks below a threshold are discarded. While effective, however, as discussed in Sec.~\ref{supple_sec:curation_details}, this method does not directly assess the reliability of intermediate frames.
In contrast, our approach defines failure at the chunk level by measuring the disagreement between forward and backward SAM2 predictions, yielding a per-frame reliability signal without requiring additional manual annotations. This enables fully automatic large-scale curation while remaining applicable beyond Ego-Exo4D.

\paragrapht{Video Diffusion Model.}
Video diffusion models have evolved from U-Net backbones inflated from image diffusion~\cite{blattmann2023stable, blattmann2023alignlatentshighresolutionvideo, ho2022videodiffusionmodels}, which effectively capture local motion but struggle to scale to long-range dependencies, to Diffusion Transformers (DiT)~\cite{peebles2023scalablediffusionmodelstransformers} that model global spatiotemporal interactions through full attention~\cite{videoworldsimulators2024, yang2024cogvideox, li2024hunyuan, polyak2025moviegencastmedia}. Building on this architectural shift, recent advances have further redefined the generative paradigm through Flow Matching~\cite{lipman2022flow, liu2022flowstraightfastlearning}, which generalizes diffusion by learning a continuous velocity field along a deterministic ODE trajectory between arbitrary distributions. This formulation has been adopted by state-of-the-art video DiTs such as Wan~\cite{wan2025wan}, enabling more flexible and efficient modeling of complex video dynamics.

Rather than using these models purely for synthesis, we repurpose their role through the lens of perception. Specifically, we build on Wan as our backbone, leveraging its pretrained spatiotemporal priors and repurposing them for cross-view reasoning. By doing so, we redirect the model’s generative capacity toward predicting object masks jointly across views and over time, enabling consistent object correspondence in a unified spatiotemporal framework.


\newpage
\clearpage

\end{document}